\documentclass{article}

\usepackage[arxiv, final]{neurips_2026}

\usepackage[utf8]{inputenc} 
\usepackage[T1]{fontenc}    
\usepackage{hyperref}       
\usepackage{url}            
\usepackage{booktabs}       
\usepackage{amsfonts}       
\usepackage{nicefrac}       
\usepackage{microtype}      
\usepackage{xcolor}         

\usepackage{amsmath}
\usepackage{amssymb}
\usepackage{amsthm}
\usepackage{booktabs}
\usepackage[nameinlink,nosort]{cleveref}
\usepackage{enumitem}
\usepackage{graphicx}
\usepackage{subcaption}
\usepackage{tabularx}

\AtBeginDocument{

\newtheorem{theorem}{Theorem}
\newtheorem{proposition}[theorem]{Proposition}
\newtheorem{lemma}[theorem]{Lemma}

\makeatletter
\let\OriginalNatbibHyper\NAT@hyper@
\renewcommand{\NAT@hyper@}[1]{\mbox{\OriginalNatbibHyper{#1}}}
\makeatother

\crefname{definition}{Def.}{Defs.}
\crefname{theorem}{Thm.}{Thms.}
\crefname{proposition}{Prop.}{Props.}
\crefname{lemma}{Lem.}{Lems.}
\crefname{corollary}{Cor.}{Cors.}
\crefname{section}{Sec.}{Secs.}
\crefname{subsection}{Sec.}{Secs.}
\crefname{subsubsection}{Sec.}{Secs.}
\crefname{appendix}{App.}{Apps.}
\crefname{subappendix}{App.}{Apps.}
\crefname{subsubappendix}{App.}{Apps.}
\AddToHook{cmd/appendix/after}{
    \crefalias{section}{appendix}
    \crefalias{subsection}{subappendix}
    \crefalias{subsubsection}{subsubappendix}
}
\crefname{figure}{Fig.}{Figs.}
\crefname{table}{Table}{Tables}
\crefname{equation}{Eq.}{Eqs.}

\crefformat{equation}{#2Eq.~#1#3}
\crefmultiformat{equation}
    {#2Eqs.~#1#3}{, #2#1#3}{, #2#1#3}{, #2#1#3}

\ExplSyntaxOn
\RenewDocumentCommand{\autoref}{s m}
  {
    \group_begin:
    \clist_set:Nn \l_tmpa_clist {#2}
    \tl_set:Nx \l_tmpa_tl {\clist_use:Nn \l_tmpa_clist {,}}
    \IfBooleanTF{#1}
      {\exp_args:NNV \cref * \l_tmpa_tl}
      {\exp_args:NV \cref \l_tmpa_tl}
    \group_end:
  }
\ExplSyntaxOff

\makeatletter
\ExplSyntaxOn
\cs_new_protected:Npn \autorefListConjunction
  {
    \group_begin:
    \cref@isstackfull{\@refstack}
    \if@cref@stackfull ,\space \else \space and\nobreakspace \fi
    \group_end:
  }
\cs_new_protected:Npn \autorefGroupConjunction
  {
    \group_begin:
    \if@cref@compress
      \exp_args:NV \cref@processconsecutive \cref@variant@get
        {\@refsubstack}{\@beginref}{\@endref}{\count@consecutive}
    \else
      \cref@stack@pop{\@refsubstack}
      \count@consecutive=1\relax
    \fi
    \cref@isstackfull{\@refsubstack}
    \ifnum\count@consecutive=2\relax
      \@cref@stackfulltrue
    \fi
    \if@cref@stackfull ,\space \else \space and\nobreakspace \fi
    \group_end:
  }
\ExplSyntaxOff
\makeatother

\crefmultiformat{equation}
    {#2Eqs.~#1#3}
    {\autorefListConjunction#2#1#3}
    {, #2#1#3}
    {\autorefListConjunction#2#1#3}}

\title{How Diffusion Models Memorize}

\author{
    Juyeop Kim \qquad
    Songkuk Kim\textsuperscript{*} \qquad
    Jong-Seok Lee\textsuperscript{*}
    \\
    Yonsei University, Korea
    \\
    \texttt{\{juyeopkim, songkuk, jong-seok.lee\}@yonsei.ac.kr}
}

\begin{document}

\maketitle

\begin{NoHyper} 
\begingroup
\renewcommand{\thefootnote}{*}
\footnotetext{Corresponding authors.}
\endgroup
\end{NoHyper}

\begin{abstract}
Despite their success in image generation, diffusion models can unintentionally memorize and reproduce training data, raising serious privacy and copyright concerns. Although prior work has studied memorization from different perspectives, a unified account of how this reproduction occurs during generation remains incomplete. In this paper, we develop a mathematical account of this process in text-to-image latent diffusion models with classifier-free guidance. We trace reproduction across denoising steps through predictions of the final denoised latent made with and without the prompt. Assuming these predictions are Bayes-optimal and uniformly bounded, we show that sufficiently low loss on a training prompt--image pair enables the prediction made with the prompt to reproduce the paired training latent with high probability, even at the first denoising step. At this first step, we bound the difference between the prediction made without the prompt and the empirical mean of the training latents, showing how guidance amplifies the reproduced training latent's deviation from that mean. We then quantify how the latent passed between denoising steps carries prompt influence to subsequent predictions made without the prompt. When both predictions are sufficiently close to the paired training latent at the final step, their guided combination also reproduces it in the final generated latent. We also analyze how their difference evolves, showing that closer agreement alone does not establish reproduction. Finally, our account provides a unified explanation of existing detection and mitigation strategies.
\end{abstract}

\section{Introduction}
\label{sec:introduction}

Text-to-image (T2I) diffusion models have emerged as a dominant paradigm for image generation~\citep{starodubcev2024invertible}, allowing users to generate high-quality images by specifying their desired content through text prompts~\citep{ramesh2022dalle2,rombach2022sdv1,nichol2022glide}.
While the models have surpassed earlier approaches in image quality and prompt--image alignment~\citep{nichol2022glide,saharia2022imagen}, they can also exhibit unintended \emph{memorization} of training prompt--image pairs, generating near-identical copies of training images in response to their paired prompts even across different random seeds~\citep{webster2023extraction,somepalli2023forgery,carlini2023extraction,wen2024detecting,hintersdorf2024nemo}.
This behavior raises serious privacy and copyright concerns, as it can expose sensitive information or reproduce copyrighted content~\citep{carlini2023extraction,jiang2023aiart}.

To address this issue, prior work has sought to understand memorization from different perspectives~\citep{somepalli2023copying,gu2025memorization,baptista2025regularization,ross2025geometric,jeon2025sharpness}.
Yet a unified account of how a model reproduces a training image from its paired prompt remains incomplete.
Without such an account, we lack a common framework for explaining the effectiveness of existing detection and mitigation strategies, which have been motivated by intuition and empirical observations of model behavior~\citep{chen2024memorizationfree,wen2024detecting,ren2024crossattention,hintersdorf2024nemo,jain2025basin}.
Their empirical success alone does not establish whether the same model behaviors remain reliable indicators of memorization or whether the same mitigation strategies remain effective beyond the settings in which they were evaluated.
Connecting these model behaviors to training-image reproduction provides a basis for assessing the reliability and limitations of existing strategies and guiding their adaptation to new models and generation settings.

In this paper, we develop a mathematical account of how memorization occurs during generation in T2I diffusion models.
Specifically, assuming Bayes-optimal and uniformly bounded model predictions, we formalize how information learned from a training prompt--image pair can lead to reproduction of the training image from its paired prompt at inference time, focusing on widely used latent diffusion models~\citep{rombach2022sdv1,li2025anydressing} with classifier-free guidance (CFG)~\citep{ho2021cfg,meng2023distillation}.
Our contributions are as follows:
\begin{enumerate}[leftmargin=*]

\item
We trace memorization through the first, subsequent, and final denoising steps using \emph{clean predictions}, which estimate the final denoised latent representation of the output image.
At the first step, we derive a lower bound on the probability that the clean prediction from CFG's \emph{conditional branch}, which receives the prompt, reproduces the paired training latent.
This bound yields a threshold on the pair's expected training loss determined by the noise schedule: \textbf{lower loss strengthens the guarantee of reproduction even at the first denoising step}.
For CFG's \emph{unconditional branch}, which receives no prompt, we bound the difference between its clean prediction and the empirical mean of the training latents using the noise schedule and dataset statistics.
Together, these findings establish that \textbf{CFG amplifies the reproduced training latent's deviation from the empirical mean by the guidance scale}, up to the derived error bounds.

\item
At subsequent denoising steps, we derive a condition for continued reproduction of the paired training latent by the conditional branch and bound how a CFG update changes the unconditional branch's clean prediction at the next step through the latent produced by that update.
This latent carries the prompt's influence from the preceding step and is supplied to both branches.
Thus, \textbf{even though the unconditional branch receives no prompt, it can be influenced by the prompt through its input}, explaining how the initial amplification can affect predictions during subsequent denoising steps.

\item
At the final denoising step, we bound the difference between the final generated latent and the paired training latent using the clean-prediction errors of both branches and the guidance scale.
This establishes that \textbf{if both branches reproduce the paired training latent sufficiently accurately, their CFG combination also reproduces it in the final generated latent}.
Together with the preceding results, this connects the pair's training loss and its influence during denoising to the conditions for final reproduction.

\item
We further examine whether agreement between the branches indicates memorization.
In an idealized model, we characterize when the gap between the probabilities the branches assign to the paired training latent widens or narrows as their shared input increasingly favors that latent.
We connect this probability gap to the norm of the difference between the branches' clean predictions and bound how contributions from other training latents affect this norm.
These results show that \textbf{closer agreement alone does not establish reproduction of the paired training latent}.

\item
Finally, we use our account to explain existing detection and mitigation strategies within a common mathematical framework.
We connect the model behaviors used for detection and the changes introduced for mitigation to how the two branches' clean predictions lead to reproduction of the paired training latent.
This provides \textbf{a unified mathematical explanation of why these strategies work}, linking prior empirical findings to the mechanisms established by our analysis.

\end{enumerate}
\section{Preliminaries}
\label{sec:preliminaries}

\textbf{Forward and reverse processes.}
Diffusion models consist of a forward process that gradually adds noise to data and a reverse process that generates samples by denoising~\citep{sohldickstein2015thermodynamics,ho2020ddpm}.
For computational efficiency, diffusion is often performed in the lower-dimensional latent space $\mathbb{R}^{d}$ of a pretrained autoencoder~\citep{rombach2022sdv1}.
Let $\beta_t\in(0,1)$ denote the noise schedule, and define $\alpha_t:=1-\beta_t$ and $\bar{\alpha}_t:=\prod_{s=1}^{t}\alpha_s$.
Given a clean latent $\mathbf{x}\sim q$, the forward process adds noise over $T$ steps, with the distribution at timestep $t$ given by
\begin{equation}
\label{eq:forward_process_closed_form}
    q(\mathbf{x}_t\mid\mathbf{x})
    =
    \mathcal{N}\!\left(
        \mathbf{x}_t;
        \sqrt{\bar{\alpha}_t}\,\mathbf{x},
        (1-\bar{\alpha}_t)\mathbf{I}
    \right).
\end{equation}
Equivalently, the forward latent $\mathbf{x}_t$ can be sampled through reparameterization as
\begin{equation}
\label{eq:forward}
    \mathbf{x}_t
    =
    \sqrt{\bar{\alpha}_t}\,\mathbf{x}
    +
    \sqrt{1-\bar{\alpha}_t}\,\boldsymbol{\epsilon},
    \qquad
    \boldsymbol{\epsilon}\sim\mathcal{N}(\mathbf{0},\mathbf{I}).
\end{equation}
The schedule decreases $\bar{\alpha}_t$ from $\bar{\alpha}_0=1$ toward zero.
The reverse process generates $\mathbf{x}_0$ by denoising $\mathbf{x}_T\sim\mathcal{N}(\mathbf{0},\mathbf{I})$.
Skipping steps can accelerate sampling~\citep{song2021ddim}.

\textbf{Training and inference.}
The denoising network is trained with the per-sample loss~\citep{ho2020ddpm}
\begin{equation}
\label{eq:loss}
    \mathcal{L}_c(\mathbf{x},t)
    :=
    \left\|
        \boldsymbol{\epsilon}
        -
        \boldsymbol{\epsilon}_c\!\left(
            \sqrt{\bar{\alpha}_t}\,\mathbf{x}
            +
            \sqrt{1-\bar{\alpha}_t}\,\boldsymbol{\epsilon},
            t
        \right)
    \right\|_2^2,
\end{equation}
where $\boldsymbol{\epsilon}_c(\mathbf{z},t)$ denotes the denoising network's noise prediction for input $\mathbf{z}$ at timestep $t$, conditioned on the text prompt $c$.
Training minimizes the expectation of this loss over training pairs, timesteps, and forward noise.
For a fixed pair $(\mathbf{x},c)$ and timestep $t$, the expected loss is $\mathbb{E}_{\boldsymbol{\epsilon}}[\mathcal{L}_c(\mathbf{x},t)]$.
During training, the prompt $c$ is randomly omitted (denoted by $\emptyset$), enabling the network to also learn the unconditional noise prediction $\boldsymbol{\epsilon}_{\emptyset}$ through the analogous loss $\mathcal{L}_{\emptyset}(\mathbf{x},t)$~\citep{ho2021cfg}.

For either the conditional branch ($b=c$) or the unconditional branch ($b=\emptyset$), the \emph{clean prediction} at timestep $t$ estimates the final denoised latent and is defined by
\begin{equation}
\label{eq:branch_clean_prediction}
    \hat{\mathbf{x}}_{t,b}(\mathbf{z})
    :=
    \frac{
        \mathbf{z}
        -\sqrt{1-\bar{\alpha}_t}\,
        \boldsymbol{\epsilon}_b(\mathbf{z},t)
    }{\sqrt{\bar{\alpha}_t}}.
\end{equation}
Below, we write $\hat{\mathbf{x}}_{t,b}:=\hat{\mathbf{x}}_{t,b}(\mathbf{x}_t)$ for notational simplicity.
At inference, classifier-free guidance (CFG)~\citep{ho2021cfg} combines the two noise predictions as
\begin{equation}
\label{eq:cfg}
    \boldsymbol{\epsilon}_g(\mathbf{x}_t,t,c)
    =
    g\,\boldsymbol{\epsilon}_c(\mathbf{x}_t,t)
    +
    (1-g)\,\boldsymbol{\epsilon}_{\emptyset}(\mathbf{x}_t,t),
\end{equation}
where $g>1$ is the guidance scale.
The corresponding \emph{guided clean prediction} is
\begin{equation}
\label{eq:predx}
    \hat{\mathbf{x}}_{t,g}
    =
    \frac{
        \mathbf{x}_t
        -
        \sqrt{1-\bar{\alpha}_t}\,
        \boldsymbol{\epsilon}_g(\mathbf{x}_t,t,c)
    }{
        \sqrt{\bar{\alpha}_t}
    }
    =
    g\,\hat{\mathbf{x}}_{t,c}
    +
    (1-g)\,\hat{\mathbf{x}}_{t,\emptyset}.
\end{equation}
The following sampling update shows how the guided clean prediction affects the latent at the next denoising step~\citep{song2021ddim}\footnote{
    While the displayed DDIM update is deterministic, the analysis extends to stochastic updates affine in the clean prediction, including DDPM~\citep{ho2020ddpm} (\autoref{fig:sdv1_ddpm}), since the added noise terms cancel when both updates use the same sampled noise.
}
\begin{equation}
\label{eq:reverse}
    \mathbf{x}_{t-1}
    =
    \sqrt{\bar{\alpha}_{t-1}}\,\hat{\mathbf{x}}_{t,g}
    +
    \sqrt{1-\bar{\alpha}_{t-1}}\,
    \boldsymbol{\epsilon}_g(\mathbf{x}_t,t,c).
\end{equation}
\section{How diffusion models memorize}
\label{sec:how_diffusion_models_memorize}

We say that a diffusion model has memorized a training latent $\mathbf{x}^\star$ (the \emph{target}) when its final generated latent $\mathbf{x}_0$ reproduces $\mathbf{x}^\star$ from its paired prompt $c$.
Since $\bar{\alpha}_0=1$, the sampling update in \autoref{eq:reverse} gives $\mathbf{x}_0=\hat{\mathbf{x}}_{1,g}$.
The final sample therefore depends on the CFG combination of the clean predictions from the conditional and unconditional branches (\autoref{eq:predx}), motivating us to examine how both branches contribute to target reproduction.\footnote{
Reproduction of the target by either branch alone does not guarantee memorization, since CFG combines both clean predictions.
}
At subsequent denoising steps, the branches' shared input $\mathbf{x}_t$ is shaped by earlier CFG updates, complicating the analysis of their clean predictions.
We therefore begin with the simplest case: the first denoising step, $t=T$, where the input $\mathbf{x}_T\sim\mathcal{N}(\mathbf{0},\mathbf{I})$ is pure Gaussian noise.
We then extend the analysis to subsequent denoising steps and final reproduction, completing our account of how memorization occurs during generation.
Let $\mathbf{X}\sim q$ denote a clean latent with mean $\boldsymbol{\mu}:=\mathbb{E}_{\mathbf{X}\sim q}[\mathbf{X}]$, and let $\mathbf{X}_t$ denote its forward latent in \autoref{eq:forward}.

\textbf{Bayes-optimal noise prediction.}
Following analyses based on optimal denoisers~\citep{kong2023information,daras2023consistent,kong2024interpretable}, we assume that both branches make Bayes-optimal noise predictions.
These predictions minimize expected squared error by taking the conditional mean of the injected noise:
\begin{equation}
\label{eq:bayes_optimal}
    \boldsymbol{\epsilon}_b(\mathbf{z},t)
    =
    \mathbb{E}_{\boldsymbol{\epsilon}}\!\left[
        \boldsymbol{\epsilon}
        \mid
        \mathbf{X}_t=\mathbf{z},t,b
    \right],
    \qquad
    b\in\{c,\emptyset\}.
\end{equation}
Under this assumption, the clean predictions in \autoref{eq:branch_clean_prediction} equal the corresponding posterior means:
\begin{equation}
\label{eq:branch_posterior_mean}
    \hat{\mathbf{x}}_{t,b}(\mathbf{z})
    =
    \mathbb{E}_{\mathbf{X}}\!\left[
        \mathbf{X}
        \mid
        \mathbf{X}_t=\mathbf{z},t,b
    \right].
\end{equation}
During sampling, each branch's clean prediction is this posterior mean evaluated at the current realized latent $\mathbf{z}=\mathbf{x}_t$.

\textbf{Uniformly bounded clean predictions.}
We further assume that the clean predictions are uniformly bounded over the timesteps and inputs considered:
\begin{equation}
\label{eq:bounded_posterior}
    \left\|
        \hat{\mathbf{x}}_{t,b}(\mathbf{z})
    \right\|_2
    \le M,
    \qquad
    b\in\{c,\emptyset\},
\end{equation}
for some finite $M$.
By \autoref{eq:branch_posterior_mean} and Jensen's inequality, this holds if $\|\mathbf{X}\|_2\le M$ almost surely.
For a finite empirical distribution over fixed training latents, $M$ can be chosen as their maximum norm.

Throughout this section and its proofs, we fix the data distribution $q$, the training pair $(\mathbf{x}^\star,c)$, the latent dimension $d$, the guidance scale $g$, and the bound $M$, unless stated otherwise.

\subsection{Initial reproduction and amplification}
\label{sec:initial_reproduction_and_amplification}

We first relate the pair's expected training loss to reproduction of the target by the conditional clean prediction at the first denoising step.
Since the loss in \autoref{eq:loss} is evaluated on forward latents, we first obtain a bound for the target's forward latent and then extend it to Gaussian initialization.

\begin{lemma}[Initial conditional reproduction; proof in \autoref{app:initial_conditional_reproduction}]
\label{lem:initial_conditional_reproduction}
Let $\mathbf{z}=\sqrt{\bar{\alpha}_T}\mathbf{x}^\star+\sqrt{1-\bar{\alpha}_T}\boldsymbol{\epsilon}$ be a forward latent of the target at $t=T$, with $\boldsymbol{\epsilon}\sim\mathcal{N}(\mathbf{0},\mathbf{I})$.
If $\mathbb{E}_{\boldsymbol{\epsilon}}[\mathcal{L}_c(\mathbf{x}^\star,T)]\le\varepsilon_T$, then, for every tolerance $r>0$,
\[
    \Pr\!\left(
        \left\|\hat{\mathbf{x}}_{T,c}(\mathbf{z})-\mathbf{x}^\star\right\|_2\le r
    \right)
    \ge
    1-\frac{1-\bar{\alpha}_T}{\bar{\alpha}_T}
    \frac{\varepsilon_T}{r^2}.
\]
The bound also holds at Gaussian initialization $\mathbf{z}=\mathbf{x}_T\sim\mathcal{N}(\mathbf{0},\mathbf{I})$ up to $o(1)$ as $\bar{\alpha}_T\to0$.
\end{lemma}

Thus, for sufficiently small $\bar{\alpha}_T$, a small $(1-\bar{\alpha}_T)\varepsilon_T/\bar{\alpha}_T$ allows target reproduction even at the first denoising step.
The unconditional branch, however, receives Gaussian noise without the paired prompt to identify the target.
Its clean prediction averages over clean latents (\autoref{eq:branch_posterior_mean}) and need not reproduce the target.
At Gaussian initialization, this posterior mean satisfies the following bound.

\begin{lemma}[Initial unconditional prediction; proof in \autoref{app:initial_unconditional_prediction}]
\label{lem:initial_unconditional_prediction}
Assume \autoref{eq:bayes_optimal} and $\|\mathbf{X}\|_2\le M$ $q$-almost surely.
For each realized initial latent $\mathbf{x}_T$, the unconditional clean prediction satisfies
\[
    \left\|\hat{\mathbf{x}}_{T,\emptyset}-\boldsymbol{\mu}\right\|_2
    \le M\left(e^{2\kappa_T}-1\right),
    \qquad
    \kappa_T:=
    \frac{
        \sqrt{\bar{\alpha}_T}M\|\mathbf{x}_T\|_2
        +\tfrac12\bar{\alpha}_TM^2
    }{1-\bar{\alpha}_T}.
\]
For fixed $\mathbf{x}_T$, this bound goes to zero as $\bar{\alpha}_T\to0$.
\end{lemma}

Thus, for each fixed $\mathbf{x}_T$, the unconditional clean prediction approaches the data mean $\boldsymbol{\mu}$ as $\bar{\alpha}_T\to0$.
Combining this bound with conditional target reproduction yields the following guarantee for the guided clean prediction.

\begin{theorem}[Initial CFG amplification; proof in \autoref{app:initial_cfg_amplification}]
\label{thm:initial_cfg_amplification}
Assume the conditions of \autoref{lem:initial_unconditional_prediction} and $\mathbb{E}_{\boldsymbol{\epsilon}}[\mathcal{L}_c(\mathbf{x}^\star,T)]\le\varepsilon_T$.
For $\mathbf{x}_T\sim\mathcal{N}(\mathbf{0},\mathbf{I})$, $g>1$, and any tolerance $r>0$,
\[
    \Pr\!\left(
        \left\|
            \hat{\mathbf{x}}_{T,g}
            -\left[\boldsymbol{\mu}+g(\mathbf{x}^\star-\boldsymbol{\mu})\right]
        \right\|_2>r
    \right)
    \le
    g^2\frac{1-\bar{\alpha}_T}{\bar{\alpha}_T}
    \frac{\varepsilon_T}{r^2}
    +o(1),
\]
as $\bar{\alpha}_T\to0$.
\end{theorem}

When the probability bound is small, the guided clean prediction is close to $\boldsymbol{\mu}+g(\mathbf{x}^\star-\boldsymbol{\mu})$ with high probability.
Thus, CFG amplifies the target's deviation from the data mean by the guidance scale $g$, up to the derived error bounds.
This guided clean prediction enters the sampling update in \autoref{eq:reverse}, shaping the latent passed to both branches at the next denoising step.
\subsection{Propagation of prompt influence}
\label{sec:propagation_of_prompt_influence}

We next examine how conditional reproduction and CFG amplification at the first denoising step can affect subsequent denoising steps.
The conditional branch continues to receive the paired prompt $c$, but its clean prediction also depends on the current latent $\mathbf{x}_t$, which has been shaped by earlier CFG updates.
Applying the algebraic relation underlying \autoref{lem:initial_conditional_reproduction} at this latent gives the following condition for continued target reproduction.

\begin{proposition}[Subsequent conditional reproduction; proof in \autoref{app:subsequent_conditional_reproduction}]
\label{prop:subsequent_conditional_reproduction}
Fix $t\in\{1,\ldots,T\}$, a realized sampling latent $\mathbf{x}_t$, and a tolerance $r\ge0$.
Then $\|\hat{\mathbf{x}}_{t,c}-\mathbf{x}^\star\|_2\le r$ if and only if
\[
    \left\|
        \frac{\mathbf{x}_t-\sqrt{\bar{\alpha}_t}\mathbf{x}^\star}
             {\sqrt{1-\bar{\alpha}_t}}
        -\boldsymbol{\epsilon}_c(\mathbf{x}_t,t)
    \right\|_2
    \le r\sqrt{\frac{\bar{\alpha}_t}{1-\bar{\alpha}_t}}.
\]
\end{proposition}

The unconditional branch, in contrast, receives no prompt.
Yet both branches receive the same latent $\mathbf{x}_t$, which has been shaped by earlier CFG updates.
Thus, the prompt can still influence the unconditional branch's clean prediction indirectly through the shared input $\mathbf{x}_t$.
To quantify this indirect influence, we examine how the conditional clean prediction used in a CFG update can affect the unconditional clean prediction at the next step.

For $t\in\{2,\ldots,T\}$, eliminating $\boldsymbol{\epsilon}_g$ from \autoref{eq:reverse} using \autoref{eq:predx} gives the CFG update below:
\begin{equation}
\label{eq:propagation_update}
    \mathbf{x}_{t-1}
    =\gamma_t\mathbf{x}_t+\eta_t\hat{\mathbf{x}}_{t,g},
\end{equation}
where $\gamma_t:=\sqrt{(1-\bar{\alpha}_{t-1})/(1-\bar{\alpha}_t)}$ and $\eta_t:=\sqrt{\bar{\alpha}_{t-1}}-\gamma_t\sqrt{\bar{\alpha}_t}>0$.
We compare it with an update using only the unconditional clean prediction at the same realized latent $\mathbf{x}_t$:
\begin{equation}
\label{eq:unconditional_update}
    \mathbf{x}_{t-1}^{\emptyset}
    :=\gamma_t\mathbf{x}_t+\eta_t\hat{\mathbf{x}}_{t,\emptyset}.
\end{equation}
The sampling coefficients $\gamma_t$ and $\eta_t$ determine the contributions of the current latent and the clean prediction, respectively.
Both updates share the same input $\mathbf{x}_t$ and therefore retain the effects of earlier CFG updates.
Their difference isolates the contribution of the conditional clean prediction at the current step.
Defining the difference between the branches' clean predictions as $\boldsymbol{\Delta}_t:=\hat{\mathbf{x}}_{t,c}-\hat{\mathbf{x}}_{t,\emptyset}$, \autoref{eq:predx} gives
\begin{equation}
\label{eq:conditional_displacement}
    \mathbf{x}_{t-1}-\mathbf{x}_{t-1}^{\emptyset}
    =g\eta_t\boldsymbol{\Delta}_t.
\end{equation}
When the conditional clean prediction exactly reproduces the target, the update difference is $g\eta_t(\mathbf{x}^\star-\hat{\mathbf{x}}_{t,\emptyset})$, showing how the reproduced target enters the next input to both branches.

We now compare the unconditional clean predictions at timestep $t-1$ for the two inputs $\mathbf{x}_{t-1}$ and $\mathbf{x}_{t-1}^{\emptyset}$ produced by the CFG and unconditional updates, respectively.
To quantify the difference between these predictions, we consider the line segment joining their inputs:
\begin{equation}
\label{eq:propagation_segment}
    \mathbf{z}_t(s):=\mathbf{x}_{t-1}^{\emptyset}+sg\eta_t\boldsymbol{\Delta}_t,
    \qquad s\in[0,1].
\end{equation}
Its endpoints are $\mathbf{z}_t(0)=\mathbf{x}_{t-1}^{\emptyset}$ and $\mathbf{z}_t(1)=\mathbf{x}_{t-1}$.
Then, the Jacobian $\mathbf{J}_{t-1,\emptyset}(\mathbf{z}):=D_{\mathbf{z}}\hat{\mathbf{x}}_{t-1,\emptyset}(\mathbf{z})$ describes how the unconditional clean prediction responds to changes in its input.
Integrating this Jacobian along the segment gives the following exact identity for the difference between the two clean predictions and a bound on its norm.

\begin{theorem}[Propagation of prompt influence; proof in \autoref{app:propagation_of_prompt_influence}]
\label{thm:propagation_of_prompt_influence}
For $t\in\{2,\ldots,T\}$, suppose $\hat{\mathbf{x}}_{t-1,\emptyset}(\cdot)$ is continuously differentiable on a neighborhood of the segment in \autoref{eq:propagation_segment}.
Then
\[
    \hat{\mathbf{x}}_{t-1,\emptyset}(\mathbf{x}_{t-1})
    -\hat{\mathbf{x}}_{t-1,\emptyset}(\mathbf{x}_{t-1}^{\emptyset})
    =g\eta_t\int_0^1
    \mathbf{J}_{t-1,\emptyset}(\mathbf{z}_t(s))
    \boldsymbol{\Delta}_t\,ds,
\]
and consequently
\[
    \left\|
        \hat{\mathbf{x}}_{t-1,\emptyset}(\mathbf{x}_{t-1})
        -\hat{\mathbf{x}}_{t-1,\emptyset}(\mathbf{x}_{t-1}^{\emptyset})
    \right\|_2
    \le g\eta_t\sup_{s\in[0,1]}
    \left\|\mathbf{J}_{t-1,\emptyset}(\mathbf{z}_t(s))\right\|_{\mathrm{op}}
    \|\boldsymbol{\Delta}_t\|_2.
\]
\end{theorem}

This result quantifies how the prompt's contribution to a CFG update can affect the next unconditional clean prediction through the latent supplied as input, even though the unconditional branch receives no prompt.
At timestep $t-1$, this unconditional clean prediction enters CFG again, so the indirect prompt influence can also affect the guided clean prediction.
\subsection{Final reproduction}
\label{sec:final_reproduction}

We now relate the clean predictions at the final denoising step to reproduction of the target in the final generated latent.
Using $\mathbf{x}_0=\hat{\mathbf{x}}_{1,g}$ and \autoref{eq:predx} at the realized latent $\mathbf{x}_1$, we obtain
\begin{equation}
\label{eq:final_error_identity}
    \mathbf{x}_0-\mathbf{x}^\star
    =\bigl(\hat{\mathbf{x}}_{1,c}-\mathbf{x}^\star\bigr)
    +(g-1)\boldsymbol{\Delta}_1,
\end{equation}
where $\boldsymbol{\Delta}_1:=\hat{\mathbf{x}}_{1,c}-\hat{\mathbf{x}}_{1,\emptyset}$.
Thus, final reproduction depends on the conditional clean prediction's error and the difference between the branches' clean predictions.
Under \autoref{eq:bayes_optimal}, both clean predictions equal the corresponding posterior means in \autoref{eq:branch_posterior_mean} evaluated at $\mathbf{x}_1$.
To quantify the target's contribution to these means, we define the \emph{target posterior probability} at timestep $t\in\{1,\ldots,T\}$ and input $\mathbf{z}\in\mathbb{R}^d$ by
\begin{equation}
\label{eq:final_target_posterior}
    \pi_{t,b}(\mathbf{z})
    :=\Pr\!\left(
        \mathbf{X}=\mathbf{x}^\star
        \mid\mathbf{X}_t=\mathbf{z},t,b
    \right),
    \qquad b\in\{c,\emptyset\}.
\end{equation}
At $t=1$ and $\mathbf{z}=\mathbf{x}_1$, the law of total expectation gives
\begin{equation}
\label{eq:final_posterior_decomposition}
    \begin{aligned}
        \hat{\mathbf{x}}_{1,b}
        &=\pi_{1,b}(\mathbf{x}_1)\mathbf{x}^\star
        \\
        &\quad+\bigl[1-\pi_{1,b}(\mathbf{x}_1)\bigr]
        \mathbb{E}_{\mathbf{X}}\!\left[
            \mathbf{X}
            \mid\mathbf{X}_1=\mathbf{x}_1,1,b,
            \mathbf{X}\ne\mathbf{x}^\star
        \right].
    \end{aligned}
\end{equation}
The second term in \autoref{eq:final_posterior_decomposition} represents the contribution of other clean latents to the clean prediction and is interpreted as zero when $\pi_{1,b}(\mathbf{x}_1)=1$, since these latents then have zero posterior probability.
The following theorem uses this decomposition and CFG to bound the final reproduction error, $\|\mathbf{x}_0-\mathbf{x}^\star\|_2$.

\begin{theorem}[Final reproduction; proof in \autoref{app:final_reproduction}]
\label{thm:final_reproduction}
Assume \autoref{eq:bayes_optimal} and $\|\mathbf{X}\|_2\le M$ almost surely under the conditional and unconditional data distributions.
For each realized latent $\mathbf{x}_1$ at the final denoising step and $g>1$, the final generated latent satisfies
\[
    \|\mathbf{x}_0-\mathbf{x}^\star\|_2
    \le\bigl(M+\|\mathbf{x}^\star\|_2\bigr)
    \Bigl[
        g\bigl(1-\pi_{1,c}(\mathbf{x}_1)\bigr)
        +(g-1)\bigl(1-\pi_{1,\emptyset}(\mathbf{x}_1)\bigr)
    \Bigr].
\]
\end{theorem}

Thus, sufficiently high target posterior probabilities in both branches ensure that the final generated latent is close to the target.
\subsection{Memorization dynamics}
\label{sec:memorization_dynamics}

Since sufficiently accurate target reproduction by both branches ensures final reproduction, one might expect earlier or closer agreement between their clean predictions to indicate stronger memorization.
We investigate this intuition by studying $\boldsymbol{\Delta}_t$ during denoising, which under \autoref{eq:bayes_optimal} is the difference between the conditional and unconditional posterior means.
We therefore call $\boldsymbol{\Delta}_t$ the \emph{posterior lag}, motivated by the possibility that the conditional clean prediction is already close to the target while the unconditional clean prediction remains farther away.
Accordingly, the \emph{posterior lag magnitude}, $\|\boldsymbol{\Delta}_t\|_2$, measures the separation between the branches' clean predictions.
This analysis clarifies how to interpret posterior lag trajectories and provides a mathematical basis for developing detection and mitigation strategies.

The paired prompt can give the target a higher prior probability than it has without the prompt.
We express this advantage through the \emph{target prior probabilities}:
\begin{equation}
\label{eq:target_prior_probabilities}
    p_b:=\Pr(\mathbf{X}=\mathbf{x}^\star\mid b),
    \qquad b\in\{c,\emptyset\},
    \qquad 0<p_{\emptyset}<p_c<1.
\end{equation}
To isolate the effect of this prior advantage on the target posterior probabilities, we consider an idealized model in which both branches use the same evidence from the current latent.
Specifically, after excluding the target, we assume that the average forward likelihood of the current latent is the same for both branches:
\begin{equation}
\label{eq:non_target_likelihood_equality}
    \mathbb{E}_{\mathbf{X}}\!\left[
        q(\mathbf{x}_t\mid\mathbf{X})
        \mid c,\mathbf{X}\ne\mathbf{x}^\star
    \right]
    =
    \mathbb{E}_{\mathbf{X}}\!\left[
        q(\mathbf{x}_t\mid\mathbf{X})
        \mid\emptyset,\mathbf{X}\ne\mathbf{x}^\star
    \right].
\end{equation}
Each expectation averages the Gaussian forward density in \autoref{eq:forward_process_closed_form} over the corresponding data distribution, excluding the target.
Since the target likelihood $q(\mathbf{x}_t\mid\mathbf{x}^\star)$ is also common to both branches, we define their shared \emph{target evidence} by
\begin{equation}
\label{eq:target_evidence}
    \lambda_t
    :=\log\frac{
        q(\mathbf{x}_t\mid\mathbf{x}^\star)
    }{
        \mathbb{E}_{\mathbf{X}}\!\left[
            q(\mathbf{x}_t\mid\mathbf{X})
            \mid\emptyset,\mathbf{X}\ne\mathbf{x}^\star
        \right]
    }.
\end{equation}
Larger $\lambda_t$ means that the current latent more strongly favors the target over other clean latents.
When $\lambda_t=0$, the target's forward likelihood equals their average forward likelihood.
Bayes' rule gives
\begin{equation}
\label{eq:target_posterior_from_evidence}
    \pi_{t,b}(\mathbf{x}_t)
    =\frac{p_b e^{\lambda_t}}{p_b e^{\lambda_t}+1-p_b}
    =\sigma\!\left(\lambda_t+\operatorname{logit}(p_b)\right),
    \qquad b\in\{c,\emptyset\},
\end{equation}
where $\sigma(u):=(1+\exp(-u))^{-1}$ and $\operatorname{logit}(p):=\log(p/(1-p))$ for $p\in(0,1)$.
For the same evidence, the larger conditional prior gives the target a higher posterior probability.
The unconditional branch therefore needs stronger evidence to reach the same target posterior probability.
To study how the difference between these probabilities changes with evidence, we define the \emph{posterior probability gap}:
\begin{equation}
\label{eq:posterior_probability_gap}
    G(\lambda)
    :=\sigma\!\left(\lambda+\operatorname{logit}(p_c)\right)
    -\sigma\!\left(\lambda+\operatorname{logit}(p_{\emptyset})\right),
\end{equation}
so $G(\lambda_t)=\pi_{t,c}(\mathbf{x}_t)-\pi_{t,\emptyset}(\mathbf{x}_t)>0$.
The following lemma characterizes $G$ as evidence increases from its first-step value $\lambda_T$, with $\lambda_T=0$ representing the ideal uninformative case.

\begin{lemma}[Posterior probability gap; proof in \autoref{app:posterior_probability_gap}]
\label{lem:posterior_probability_gap}
Assume $0<p_{\emptyset}<p_c<1$.
The posterior probability gap $G(\lambda)$ is unimodal in the target evidence $\lambda$, with unique maximizer
\[
    \lambda^\star
    :=-\frac{\operatorname{logit}(p_c)+\operatorname{logit}(p_{\emptyset})}{2}.
\]
On $[\lambda_T,\infty)$, $G$ strictly decreases if $\lambda_T\ge\lambda^\star$.
Otherwise, it strictly increases on $[\lambda_T,\lambda^\star]$ and strictly decreases on $[\lambda^\star,\infty)$.
When $\lambda_T=0$, these cases correspond to $p_c+p_{\emptyset}\ge1$ and $p_c+p_{\emptyset}<1$, respectively.
\end{lemma}

If target evidence increases during denoising, both branches assign increasing probability to the target.
Below $\lambda^\star$, the conditional probability increases faster, widening the gap.
Above $\lambda^\star$, the unconditional probability increases faster and catches up, narrowing the gap.
The following theorem connects this probability gap to posterior lag by bounding the difference between $\|\boldsymbol{\Delta}_t\|_2$ and $G(\lambda_t)\|\mathbf{x}^\star\|_2$ in terms of contributions from other clean latents.

\begin{theorem}[Posterior lag dynamics; proof in \autoref{app:posterior_lag_dynamics}]
\label{thm:posterior_lag_dynamics}
Assume \autoref{eq:bayes_optimal} and $0<p_{\emptyset}<p_c<1$.
For each timestep $t\in\{1,\ldots,T\}$ and realized latent $\mathbf{x}_t$ satisfying \autoref{eq:non_target_likelihood_equality}, the posterior lag and its magnitude satisfy
\[
    \begin{aligned}
        \left|
            \|\boldsymbol{\Delta}_t\|_2
            -G(\lambda_t)\|\mathbf{x}^\star\|_2
        \right|
        &\le
        \|\boldsymbol{\Delta}_t-G(\lambda_t)\mathbf{x}^\star\|_2
        \\
        &\le
        \sum_{b\in\{c,\emptyset\}}
        \bigl[1-\pi_{t,b}(\mathbf{x}_t)\bigr]
        \left\|
            \mathbb{E}_{\mathbf{X}}\!\left[
                \mathbf{X}
                \mid\mathbf{X}_t=\mathbf{x}_t,t,b,
                \mathbf{X}\ne\mathbf{x}^\star
            \right]
        \right\|_2.
    \end{aligned}
\]
\end{theorem}

Therefore, small weighted contributions from other clean latents keep the posterior lag magnitude close to the ideal curve $G(\lambda_t)\|\mathbf{x}^\star\|_2$.
Since $G(\lambda)\to0$ as $\lambda\to\pm\infty$, the magnitude can be small when both branches assign either little or nearly all probability to the target.
Thus, close agreement alone does not establish target reproduction: the predictions may agree away from the target when target evidence is very low.
To understand where they agree, we must account for the target's posterior probability in each branch, which determines its weight in the prediction, and the remaining contributions from other clean latents.
\section{Experiments}
\label{sec:experiments}

\subsection{Settings}
\label{sec:settings}

We conduct experiments with Stable Diffusion v1.4 (SDv1)~\citep{rombach2022sdv1}, Stable Diffusion v2.1 (SDv2)~\citep{lopez2022sdv2}, and RealisticVision (RealVis)~\citep{sg1612222023realvis}.
We use DDIM~\citep{song2021ddim} for all three models and additionally evaluate SDv1 with DDPM~\citep{ho2020ddpm}.
All configurations use $T=50$ inference steps and guidance scale $g=7.5$.
We present SDv1 (DDIM) results in the main text and report results for the other configurations in \autoref{app:additional_results}.
We use all available paired training images and prompts from \citet{webster2023extraction}, yielding 446 pairs for SDv1 (see \autoref{app:dataset} for details).
For each prompt, we generate 20 RGB images at a resolution of $512\times512$ pixels from Gaussian initializations $\mathbf{x}_T\sim\mathcal{N}(\mathbf{0},\mathbf{I})$.
The same 20 seeds are reused across prompts.
All experiments are run on $4\times$ NVIDIA RTX PRO 6000 Blackwell Server Edition GPUs, requiring less than $1.2$TB of storage and completing within $12$ hours.

\textbf{Quantifying memorization.}
We assess reproduction of the paired training image using SSCD~\citep{pizzi2022sscd}, which has shown strong performance in detecting image replication~\citep{somepalli2023forgery}.
We compute the cosine similarity between the SSCD descriptors of each generated image and its paired training image.
Following \citet{wen2024detecting}, we use a threshold of $0.7$.
For each model--sampler configuration, we classify a prompt as memorized if at least one of its 20 generated images has SSCD $>0.7$, and as normal if all scores are $\leq0.7$.
For SDv1 (DDIM), these groups contain 135 and 311 prompts, respectively.
All generated samples remain in their prompt's group, and prompt-group summaries give each prompt equal weight.

\textbf{Measurements.}
We encode each paired training image with the model's pretrained autoencoder to obtain the target latent $\mathbf{x}^\star$ and record both branches' clean predictions at every denoising step.
Their difference gives the posterior lag $\boldsymbol{\Delta}_t$.
Displayed Euclidean norms are divided by $\sqrt{d}$, where $d$ is the latent dimension.
To compare initial conditional reproduction under forward noise and Gaussian initialization, we additionally evaluate 20 independently sampled forward latents per target at the noise level of the first denoising step.
For measurements involving the data mean $\boldsymbol{\mu}$, we use a model-based estimate $\hat{\boldsymbol{\mu}}$, motivated by \autoref{eq:branch_posterior_mean,lem:initial_unconditional_prediction}.
For each model, we obtain this estimate by averaging 4,096 unconditional clean predictions evaluated on standard Gaussian inputs at the largest timestep used in training.
The same estimate is used for the unconditional-prediction and initial CFG measurements.
To measure propagation of prompt influence, we compare the CFG and unconditional updates from the same current latent, as in \autoref{eq:propagation_update,eq:unconditional_update}.
We hold any sampled noise fixed across the two updates and evaluate the next unconditional clean prediction at both resulting latents.
We approximate the Jacobian line integral in \autoref{thm:propagation_of_prompt_influence} using Gauss--Legendre quadrature of Jacobian--vector products.
\subsection{Results}
\label{sec:results}

\begin{figure}[t]
    \centering
    \captionsetup[subfigure]{skip=2pt,hypcap=true}
    \setlength{\parskip}{0pt}

    \begin{subfigure}[t]{0.32\linewidth}
        \centering
        \includegraphics[
            width=\linewidth,
            viewport=-6 -7 351.78 327.464,
            clip
        ]{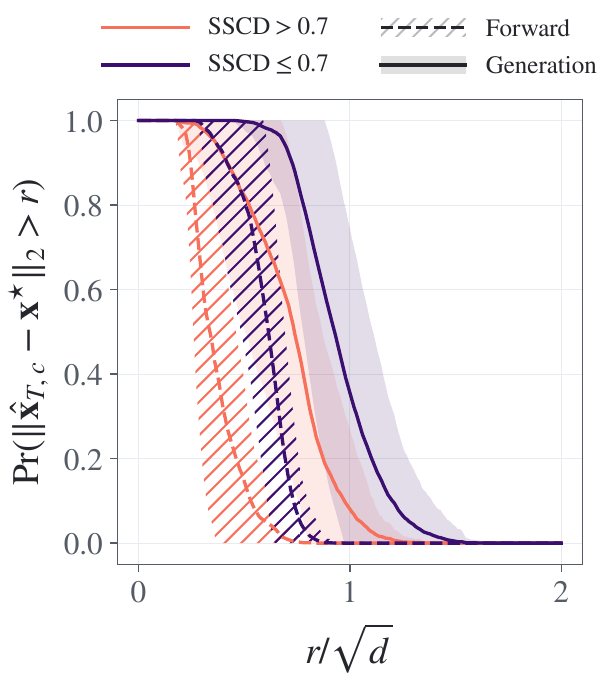}
        \caption{\autoref{lem:initial_conditional_reproduction}}
        \label{fig:sdv1_ddim_lem1}
    \end{subfigure}\hfill
    \begin{subfigure}[t]{0.32\linewidth}
        \centering
        \includegraphics[
            width=\linewidth,
            viewport=-2 -10.25 355.78 324.214,
            clip
        ]{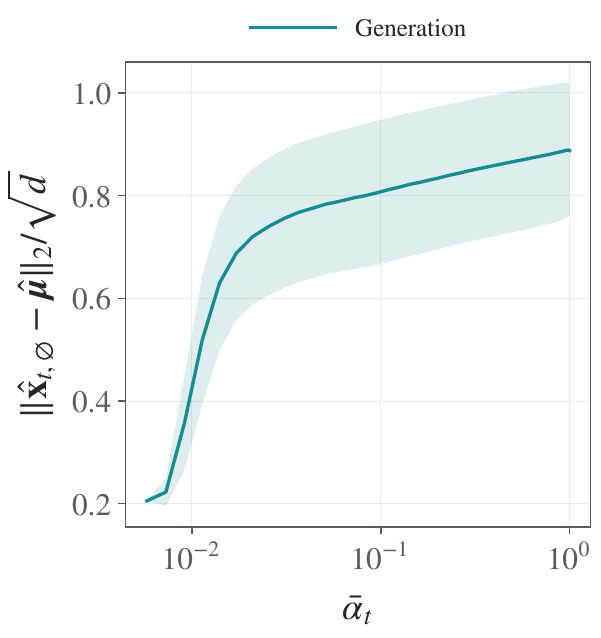}
        \caption{\autoref{lem:initial_unconditional_prediction}}
        \label{fig:sdv1_ddim_lem2}
    \end{subfigure}\hfill
    \begin{subfigure}[t]{0.32\linewidth}
        \centering
        \includegraphics[
            width=\linewidth,
            viewport=-5 -7 352.78 327.464,
            clip
        ]{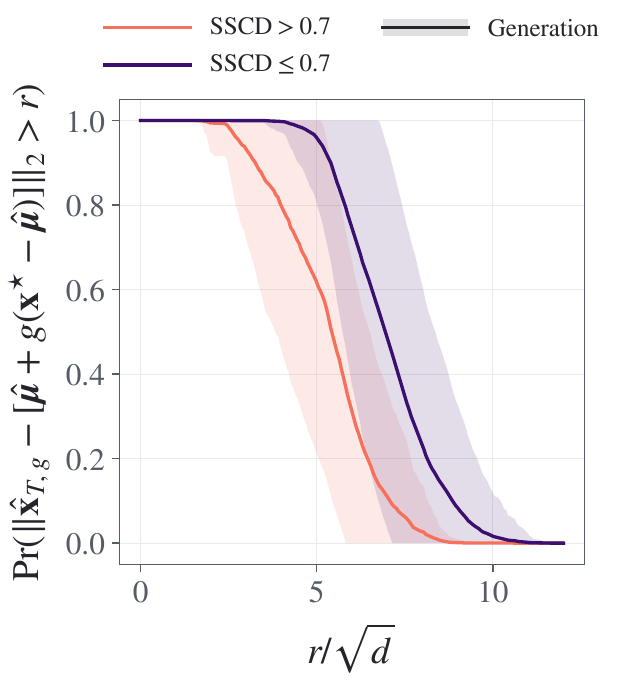}
        \caption{\autoref{thm:initial_cfg_amplification}}
        \label{fig:sdv1_ddim_thm3}
    \end{subfigure}

    \par\vspace{6pt}

    \begin{subfigure}[t]{0.32\linewidth}
        \centering
        \includegraphics[
            width=\linewidth,
            viewport=-2 -13 355.78 321.464,
            clip
        ]{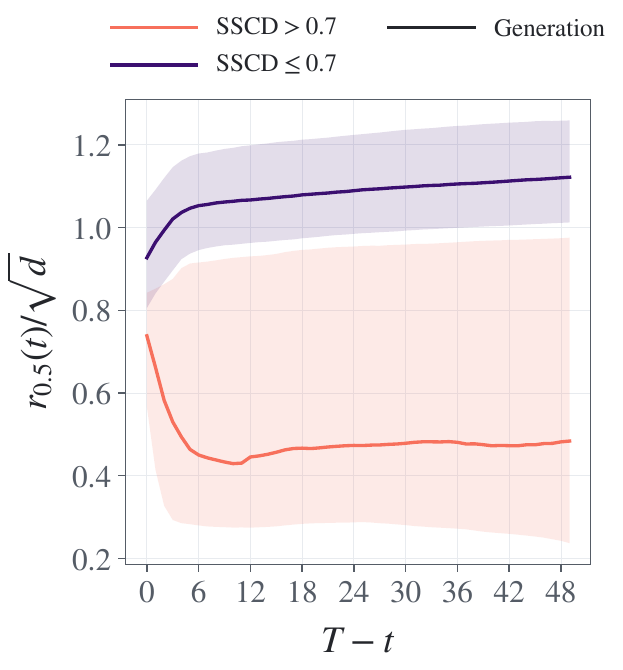}
        \caption{\autoref{prop:subsequent_conditional_reproduction}}
        \label{fig:sdv1_ddim_prop4}
    \end{subfigure}\hfill
    \begin{subfigure}[t]{0.32\linewidth}
        \centering
        \includegraphics[
            width=\linewidth,
            viewport=0 0 357.78 334.464,
            clip
        ]{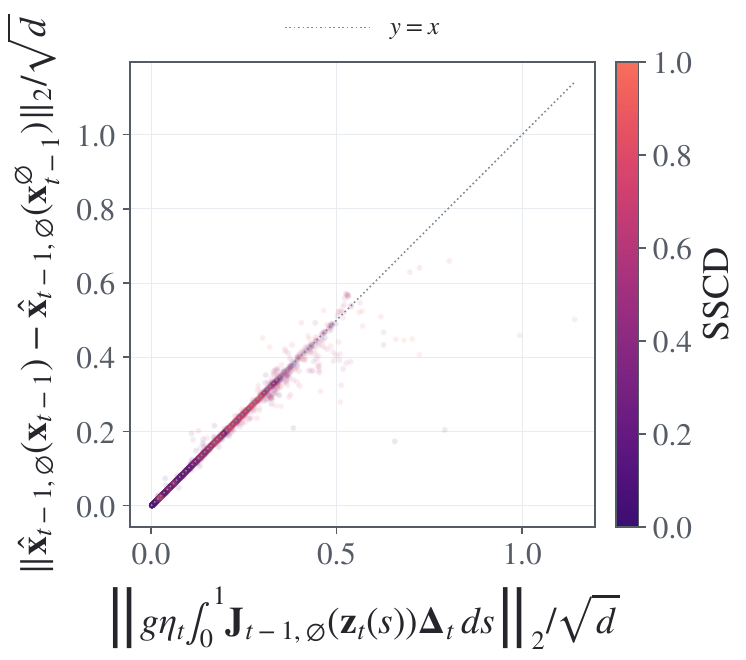}
        \caption{\autoref{thm:propagation_of_prompt_influence}}
        \label{fig:sdv1_ddim_thm5}
    \end{subfigure}\hfill
    \begin{subfigure}[t]{0.32\linewidth}
        \centering
        \includegraphics[
            width=\linewidth,
            viewport=-2 -13 355.78 321.464,
            clip
        ]{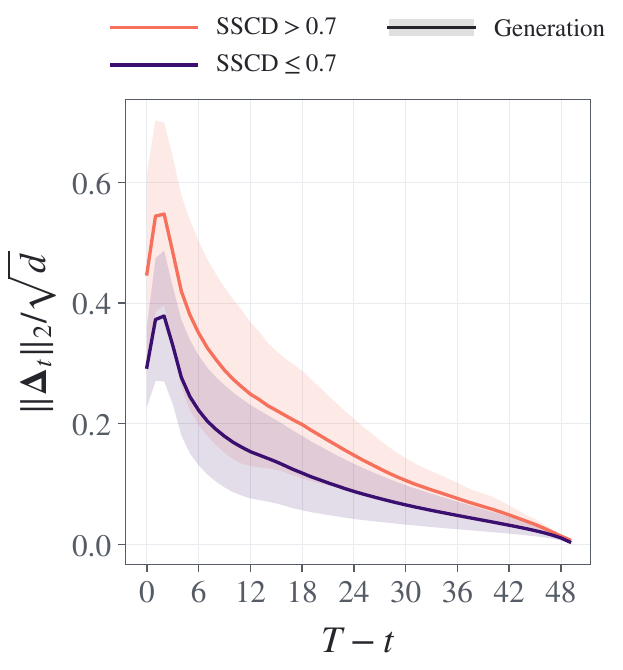}
        \caption{\autoref{thm:posterior_lag_dynamics}}
        \label{fig:sdv1_ddim_thm8}
    \end{subfigure}

    \caption{
        \textbf{Denoising measurements on SDv1 (DDIM).}
        Coral and purple in (a,c,d,f) denote memorized and normal prompt groups, respectively (\autoref{sec:settings}).
        In (a), dashed and solid curves correspond to forward latents and Gaussian initializations, respectively.
        In (a--c,f), curves show means across prompts, and shaded bands indicate $\pm1$ standard deviation.
        Seeds are first averaged within each prompt in (b,f).
        In (d), curves and shaded bands show the median and interquartile range, respectively, with each prompt receiving equal total weight.
        In (e), point colors indicate each generated image's final SSCD, and the dotted diagonal denotes equality.
    }
    \label{fig:sdv1_ddim}
\end{figure}

\textbf{Initial conditional reproduction.}
\autoref{fig:sdv1_ddim_lem1} reports the probability that the initial conditional clean prediction lies farther than a tolerance $r$ from the target.
For both forward latents and Gaussian initializations, memorized prompts have lower failure probabilities than normal prompts at the same tolerance, indicating closer initial conditional predictions to their targets (coral curves below the corresponding purple curves).
Gaussian initializations require larger tolerances than forward latents to achieve comparable failure probabilities (solid curves shifted to the right of the corresponding dashed curves).
This separation is consistent with the finite-noise distribution mismatch in \autoref{lem:initial_conditional_reproduction}: forward latents retain the target-dependent component $\sqrt{\bar{\alpha}_T}\mathbf{x}^\star$ and have noise variance $1-\bar{\alpha}_T$, whereas Gaussian initializations have zero mean and unit variance.
The two input distributions approach each other as $\bar{\alpha}_T\to0$.

\textbf{Initial unconditional prediction.}
\autoref{fig:sdv1_ddim_lem2} plots the distance between the unconditional clean prediction and the estimated data mean $\hat{\boldsymbol{\mu}}$ against $\bar{\alpha}_t$ along the generated trajectories.
The distance is smallest at the first denoising steps, where $\bar{\alpha}_t$ is smallest, and increases as denoising proceeds (teal curve rising from left to right).
In \autoref{lem:initial_unconditional_prediction}, decreasing $\bar{\alpha}_T$ at fixed $\mathbf{x}_T$ decreases the bound on the distance to the data mean, with the bound going to zero as $\bar{\alpha}_T\to0$.
The observed proximity to $\hat{\boldsymbol{\mu}}$ at small $\bar{\alpha}_t$ agrees with the initial unconditional behavior characterized by the lemma.

\textbf{Initial CFG amplification.}
\autoref{fig:sdv1_ddim_thm3} measures the probability that the initial guided clean prediction deviates by more than $r$ from $\hat{\boldsymbol{\mu}}+g(\mathbf{x}^\star-\hat{\boldsymbol{\mu}})$.
Memorized prompts have lower failure probabilities at the same tolerance, indicating closer agreement with this extrapolated target (coral curve shifted to the left of the purple curve).
This agrees with \autoref{thm:initial_cfg_amplification}: when the conditional clean prediction is close to the target and the unconditional clean prediction is close to the data mean, CFG amplifies the target's deviation from that mean by $g$.
The stronger agreement for memorized prompts connects their initial conditional reproduction to the predicted CFG amplification.

\textbf{Subsequent conditional reproduction.}
\autoref{fig:sdv1_ddim_prop4} tracks the median reproduction tolerance $r_{0.5}(t)$ under equal prompt weighting.
A smaller tolerance means that at least half of the samples under this weighting have conditional clean predictions within a smaller distance of their paired targets.
For memorized prompts, the tolerance decreases sharply during the first few updates and remains substantially below that of normal prompts throughout denoising (coral curve).
Normal prompts instead exhibit an increasing median tolerance (purple curve).
By \autoref{prop:subsequent_conditional_reproduction}, this lower median error corresponds to a smaller median discrepancy in the noise-prediction condition for target reproduction at the same timestep.
This separation shows that memorized prompts retain more accurate conditional target reproduction throughout denoising.

\textbf{Propagation of prompt influence.}
\autoref{fig:sdv1_ddim_thm5} compares the norm of the numerically integrated Jacobian response with the norm of the matched change in the next unconditional clean prediction.
Most points concentrate near the dotted equality line, showing agreement between the two quantities in \autoref{thm:propagation_of_prompt_influence}.
Larger responses tend to be associated with higher final SSCD, indicating stronger prompt influence on the next unconditional clean prediction along trajectories that more closely reproduce the target (coral points extending farther from the origin).
This matched comparison supports the propagation mechanism: the conditional contribution changes the sampling latent, which in turn changes the unconditional clean prediction at the next step.

\textbf{Memorization dynamics.}
\autoref{fig:sdv1_ddim_thm8} shows the mean posterior lag magnitude $\|\boldsymbol{\Delta}_t\|_2$ for memorized and normal prompts throughout denoising.
Both mean trajectories exhibit a single early peak followed by a gradual decrease, with memorized prompts maintaining a larger separation between the branches over most of the trajectory (coral curve above the purple curve).
The larger lag for memorized prompts is consistent with their more accurate conditional target reproduction (\autoref{fig:sdv1_ddim_lem1,fig:sdv1_ddim_prop4}), while the unconditional clean prediction initially remains near the data mean (\autoref{fig:sdv1_ddim_lem2}).
The paired prompt supplies information about the target directly to the conditional branch, whereas its influence reaches the unconditional branch through the guided sampling latent (\autoref{thm:propagation_of_prompt_influence}).
These observations are consistent with conditional target reproduction preceding comparable unconditional target reproduction, producing a larger separation between the two clean predictions.
The single-peaked mean trajectories are qualitatively consistent with the unimodal posterior probability gap in \autoref{lem:posterior_probability_gap} and its contribution to the posterior lag in \autoref{thm:posterior_lag_dynamics}.
Under the shared-evidence model, when target evidence increases through the maximizer of this gap and the weighted contributions from other clean latents are small, the conditional and unconditional target posterior probabilities first move farther apart and then closer together.
The smaller posterior lag for normal prompts, despite their larger conditional reproduction errors, also shows that closer agreement alone does not establish target reproduction.
\section{Explaining prior work}
\label{sec:explaining_prior_work}

Several prior studies propose detection or mitigation strategies for memorization in diffusion models.
Our account provides a unified mathematical explanation by relating these methods to conditional target reproduction, CFG amplification, and propagation of prompt influence.
A broader discussion of related work appears in \autoref{app:related_work}.

\textbf{\citet{wen2024detecting}} empirically show that text prompts strongly guide the denoising process under memorization and use the magnitude of the conditional--unconditional noise-prediction difference for detection and mitigation.
However, they motivate this association through intuition and empirical observations, without deriving how the prompt's contribution propagates through the sampling latent to the unconditional branch.
By \autoref{eq:branch_clean_prediction},
\begin{equation}
\label{eq:wen_noise_difference}
    \boldsymbol{\Delta}_t
    =-\sqrt{\frac{1-\bar{\alpha}_t}{\bar{\alpha}_t}}
    \left(
        \boldsymbol{\epsilon}_c(\mathbf{x}_t,t)
        -\boldsymbol{\epsilon}_{\emptyset}(\mathbf{x}_t,t)
    \right).
\end{equation}
Thus, at a fixed timestep, this noise-difference magnitude and the posterior lag magnitude in our analysis give the same ranking.
The posterior lag also determines the difference $g\eta_t\boldsymbol{\Delta}_t$ between the CFG and unconditional updates (\autoref{eq:conditional_displacement}).
\autoref{thm:propagation_of_prompt_influence} quantifies how this update difference affects the next unconditional clean prediction.
Their inference-time mitigation therefore acts on the signal that drives prompt influence through the sampling update.

\textbf{\citet{hintersdorf2024nemo}} empirically find that the first-step residuals $\boldsymbol{\epsilon}_c(\mathbf{x}_T,T)-\mathbf{x}_T$ are more consistent across seeds after normalization for memorized prompts, and use this property to identify and deactivate cross-attention neurons that trigger replication.
However, they motivate this consistency through intuition about similar denoising trajectories, without deriving why the first-step residuals share target-dependent structure across seeds.
Our analysis connects the pair's expected training loss to initial conditional reproduction (\autoref{lem:initial_conditional_reproduction}).
Rearranging \autoref{eq:branch_clean_prediction} gives
\begin{equation}
\label{eq:nemo_initial_residual}
    \begin{aligned}
        \boldsymbol{\epsilon}_c(\mathbf{x}_T,T)-\mathbf{x}_T
        &=
        \left(\frac{1}{\sqrt{1-\bar{\alpha}_T}}-1\right)\mathbf{x}_T
        \\
        &\quad-
        \sqrt{\frac{\bar{\alpha}_T}{1-\bar{\alpha}_T}}\,
        \hat{\mathbf{x}}_{T,c}.
    \end{aligned}
\end{equation}
When $\hat{\mathbf{x}}_{T,c}\approx\mathbf{x}^\star$ across seeds, the second term supplies an approximately common target-dependent component.
This explains how consistent normalized residuals can arise when that component dominates the remaining variation.
When the unconditional clean prediction is also close to the data mean, CFG approximately amplifies the reproduced target's deviation from that mean (\autoref{thm:initial_cfg_amplification}).
Their neuron-level intervention can thus be interpreted as disrupting the conditional target reproduction that supplies the target component to guided generation.

\textbf{\citet{jain2025basin}} propose mitigation by identifying a transition timestep in denoising: generation is initially unconditional, and CFG is enabled afterward.
However, their attraction-basin account does not explicitly derive how the conditional contribution to a CFG update affects the next unconditional clean prediction.
When the conditional branch reproduces the target at the first denoising step, CFG can amplify its deviation from the data mean (\autoref{thm:initial_cfg_amplification}).
At the same current latent, using an unconditional update instead removes the update difference $g\eta_t\boldsymbol{\Delta}_t$ (\autoref{eq:conditional_displacement}) and its contribution to the next unconditional clean prediction (\autoref{thm:propagation_of_prompt_influence}).
\section{Conclusion}
\label{sec:conclusion}

In this paper, we provided a theoretical explanation of how memorization occurs in text-to-image diffusion models and supported it empirically.
We showed that accurate initial conditional target reproduction and an unconditional clean prediction near the data mean allow CFG to approximately amplify the target's deviation from that mean.
We derived an exact identity describing how the prompt's contribution to a CFG update affects the next unconditional clean prediction.
Under Bayes-optimal noise prediction and bounded clean latents, we bounded final reproduction error using both branches' target posterior probabilities.
Our posterior lag analysis relates branch separation to target evidence and contributions from other clean latents.
Together, these results explain prior empirical findings and provide a theoretical basis for existing detection and mitigation strategies.
Limitations and future work are discussed in \autoref{app:limitations_and_future_work}.

\bibliography{ref}
\bibliographystyle{unsrtnat}


\clearpage
\appendix

\section{Proofs}
\label{app:proofs}

\setcounter{theorem}{0}
\renewcommand{\theHtheorem}{appendix.\arabic{theorem}}

\subsection{Proof of \autoref{lem:initial_conditional_reproduction}}
\label{app:initial_conditional_reproduction}

\begin{lemma}[Initial conditional reproduction]
Let $\mathbf{z}=\sqrt{\bar{\alpha}_T}\mathbf{x}^\star+\sqrt{1-\bar{\alpha}_T}\boldsymbol{\epsilon}$ be a forward latent of the target at $t=T$, with $\boldsymbol{\epsilon}\sim\mathcal{N}(\mathbf{0},\mathbf{I})$.
If $\mathbb{E}_{\boldsymbol{\epsilon}}[\mathcal{L}_c(\mathbf{x}^\star,T)]\le\varepsilon_T$, then, for every tolerance $r>0$,
\[
    \Pr\!\left(
        \left\|\hat{\mathbf{x}}_{T,c}(\mathbf{z})-\mathbf{x}^\star\right\|_2\le r
    \right)
    \ge
    1-\frac{1-\bar{\alpha}_T}{\bar{\alpha}_T}
    \frac{\varepsilon_T}{r^2}.
\]
The bound also holds at Gaussian initialization $\mathbf{z}=\mathbf{x}_T\sim\mathcal{N}(\mathbf{0},\mathbf{I})$ up to $o(1)$ as $\bar{\alpha}_T\to0$.
\end{lemma}

\begin{proof}
For any timestep $t\in\{1,\ldots,T\}$, substituting the forward latent $\mathbf{z}=\sqrt{\bar{\alpha}_t}\mathbf{x}^\star+\sqrt{1-\bar{\alpha}_t}\boldsymbol{\epsilon}$ into \autoref{eq:branch_clean_prediction} gives
\begin{equation}
\label{eq:conditional_reproduction_error_identity}
    \hat{\mathbf{x}}_{t,c}(\mathbf{z})-\mathbf{x}^\star
    =
    \sqrt{\frac{1-\bar{\alpha}_t}{\bar{\alpha}_t}}
    \left[
        \boldsymbol{\epsilon}
        -\boldsymbol{\epsilon}_c(\mathbf{z},t)
    \right].
\end{equation}
Taking squared norms and using \autoref{eq:loss} yields the pointwise identity
\begin{equation}
\label{eq:conditional_reproduction_loss_identity}
    \left\|\hat{\mathbf{x}}_{t,c}(\mathbf{z})-\mathbf{x}^\star\right\|_2^2
    =
    \frac{1-\bar{\alpha}_t}{\bar{\alpha}_t}
    \mathcal{L}_c(\mathbf{x}^\star,t).
\end{equation}
This identity is algebraic and does not require Bayes-optimal noise prediction.
At $t=T$, taking expectation over the forward noise gives
\begin{equation}
\label{eq:initial_conditional_reproduction_mse}
    \begin{aligned}
        \mathbb{E}_{\boldsymbol{\epsilon}}\!\left[
            \left\|\hat{\mathbf{x}}_{T,c}(\mathbf{z})-\mathbf{x}^\star\right\|_2^2
        \right]
        &=
        \frac{1-\bar{\alpha}_T}{\bar{\alpha}_T}
        \mathbb{E}_{\boldsymbol{\epsilon}}[\mathcal{L}_c(\mathbf{x}^\star,T)]
        \\
        &\le
        \frac{1-\bar{\alpha}_T}{\bar{\alpha}_T}\varepsilon_T.
    \end{aligned}
\end{equation}
Markov's inequality therefore implies that, for every $r>0$,
\begin{equation}
\label{eq:initial_conditional_reproduction_forward}
    \Pr\!\left(
        \left\|\hat{\mathbf{x}}_{T,c}(\mathbf{z})-\mathbf{x}^\star\right\|_2>r
    \right)
    \le
    \frac{1-\bar{\alpha}_T}{\bar{\alpha}_T}
    \frac{\varepsilon_T}{r^2}.
\end{equation}

To obtain the initialization bound, we compare the target's forward distribution $\mathcal{N}(\sqrt{\bar{\alpha}_T}\mathbf{x}^\star,(1-\bar{\alpha}_T)\mathbf{I})$ with the Gaussian initialization distribution $\mathcal{N}(\mathbf{0},\mathbf{I})$.
The Gaussian relative entropy is
\begin{equation}
\label{eq:initial_conditional_reproduction_kl}
    \begin{aligned}
        &D_{\mathrm{KL}}\!\left(
            \mathcal{N}\!\left(
                \sqrt{\bar{\alpha}_T}\mathbf{x}^\star,
                (1-\bar{\alpha}_T)\mathbf{I}
            \right)
            \,\middle\|\,
            \mathcal{N}(\mathbf{0},\mathbf{I})
        \right)
        \\
        &\qquad=
        \frac12\left[
            d(1-\bar{\alpha}_T)
            +\bar{\alpha}_T\|\mathbf{x}^\star\|_2^2
            -d-d\log(1-\bar{\alpha}_T)
        \right]
        \\
        &\qquad=
        \frac12\left[
            \bar{\alpha}_T\|\mathbf{x}^\star\|_2^2
            +d\left[-\bar{\alpha}_T-\log(1-\bar{\alpha}_T)\right]
        \right].
    \end{aligned}
\end{equation}
For the total variation distance, Pinsker's inequality gives
\begin{equation}
\label{eq:initial_conditional_reproduction_tv}
    \begin{aligned}
        &\operatorname{TV}\!\left(
            \mathcal{N}\!\left(
                \sqrt{\bar{\alpha}_T}\mathbf{x}^\star,
                (1-\bar{\alpha}_T)\mathbf{I}
            \right),
            \mathcal{N}(\mathbf{0},\mathbf{I})
        \right)
        \\
        &\qquad\le
        \sqrt{
            \frac12 D_{\mathrm{KL}}\!\left(
                \mathcal{N}\!\left(
                    \sqrt{\bar{\alpha}_T}\mathbf{x}^\star,
                    (1-\bar{\alpha}_T)\mathbf{I}
                \right)
                \,\middle\|\,
                \mathcal{N}(\mathbf{0},\mathbf{I})
            \right)
        }
        \\
        &\qquad=
        \frac12\sqrt{
            \bar{\alpha}_T\|\mathbf{x}^\star\|_2^2
            +d\left[-\bar{\alpha}_T-\log(1-\bar{\alpha}_T)\right]
        }.
    \end{aligned}
\end{equation}
Total variation bounds the absolute difference between the probabilities of the same event under the two distributions.
Applying this to the reproduction-error event and using \autoref{eq:initial_conditional_reproduction_forward} yields
\begin{equation}
\label{eq:initial_conditional_reproduction_gaussian}
    \begin{aligned}
        &\Pr\!\left(
            \left\|\hat{\mathbf{x}}_{T,c}-\mathbf{x}^\star\right\|_2>r
        \right)
        \\
        &\qquad\le
        \Pr\!\left(
            \left\|\hat{\mathbf{x}}_{T,c}(\mathbf{z})-\mathbf{x}^\star\right\|_2>r
        \right)
        \\
        &\qquad\quad+
        \frac12\sqrt{
            \bar{\alpha}_T\|\mathbf{x}^\star\|_2^2
            +d\left[-\bar{\alpha}_T-\log(1-\bar{\alpha}_T)\right]
        }
        \\
        &\qquad\le
        \frac{1-\bar{\alpha}_T}{\bar{\alpha}_T}
        \frac{\varepsilon_T}{r^2}
        +\frac12\sqrt{
            \bar{\alpha}_T\|\mathbf{x}^\star\|_2^2
            +d\left[-\bar{\alpha}_T-\log(1-\bar{\alpha}_T)\right]
        }
        \\
        &\qquad=
        \frac{1-\bar{\alpha}_T}{\bar{\alpha}_T}
        \frac{\varepsilon_T}{r^2}
        +o(1),
    \end{aligned}
\end{equation}
as $\bar{\alpha}_T\to0$.
Taking complements in \autoref{eq:initial_conditional_reproduction_forward,eq:initial_conditional_reproduction_gaussian} establishes the stated lower bounds on reproduction probability.
\end{proof}
\subsection{Proof of \autoref{lem:initial_unconditional_prediction}}
\label{app:initial_unconditional_prediction}

\begin{lemma}[Initial unconditional prediction]
Assume \autoref{eq:bayes_optimal} and $\|\mathbf{X}\|_2\le M$ $q$-almost surely.
For each realized initial latent $\mathbf{x}_T$, the unconditional clean prediction satisfies
\[
    \left\|\hat{\mathbf{x}}_{T,\emptyset}-\boldsymbol{\mu}\right\|_2
    \le M\left(e^{2\kappa_T}-1\right),
    \qquad
    \kappa_T:=
    \frac{
        \sqrt{\bar{\alpha}_T}M\|\mathbf{x}_T\|_2
        +\tfrac12\bar{\alpha}_TM^2
    }{1-\bar{\alpha}_T}.
\]
For fixed $\mathbf{x}_T$, this bound goes to zero as $\bar{\alpha}_T\to0$.
\end{lemma}

\begin{proof}
Fix the realized initial latent $\mathbf{x}_T$.
Expanding the forward likelihood in \autoref{eq:forward_process_closed_form} and omitting factors independent of $\mathbf{x}$ gives
\begin{equation}
\label{eq:initial_unconditional_likelihood}
    \begin{aligned}
        q(\mathbf{x}_T\mid\mathbf{x})
        &\propto
        \exp\!\left(
            -\frac{
                \|\mathbf{x}_T-\sqrt{\bar{\alpha}_T}\mathbf{x}\|_2^2
            }{2(1-\bar{\alpha}_T)}
        \right)
        \\
        &\propto
        \exp\!\left(
            \frac{
                \sqrt{\bar{\alpha}_T}\mathbf{x}_T^\top\mathbf{x}
                -\tfrac12\bar{\alpha}_T\|\mathbf{x}\|_2^2
            }{1-\bar{\alpha}_T}
        \right)
        =:w_T(\mathbf{x}).
    \end{aligned}
\end{equation}
Under Bayes-optimal noise prediction, \autoref{eq:branch_posterior_mean} and Bayes' rule yield
\begin{equation}
\label{eq:initial_unconditional_weighted_mean}
    \begin{aligned}
        Z_T
        &:=
        \mathbb{E}_{\mathbf{X}\sim q}[w_T(\mathbf{X})],
        \\
        \hat{\mathbf{x}}_{T,\emptyset}
        &=
        \mathbb{E}_{\mathbf{X}}\!\left[
            \mathbf{X}\mid\mathbf{X}_T=\mathbf{x}_T,T,\emptyset
        \right]
        =
        \frac{
            \mathbb{E}_{\mathbf{X}\sim q}[\mathbf{X}w_T(\mathbf{X})]
        }{Z_T}.
    \end{aligned}
\end{equation}
By Cauchy--Schwarz and $\|\mathbf{X}\|_2\le M$ $q$-almost surely,
\begin{equation}
\label{eq:initial_unconditional_exponent_bound}
    \begin{aligned}
        |\log w_T(\mathbf{X})|
        &\le
        \frac{
            \sqrt{\bar{\alpha}_T}\|\mathbf{x}_T\|_2\|\mathbf{X}\|_2
            +\tfrac12\bar{\alpha}_T\|\mathbf{X}\|_2^2
        }{1-\bar{\alpha}_T}
        \\
        &\le
        \frac{
            \sqrt{\bar{\alpha}_T}M\|\mathbf{x}_T\|_2
            +\tfrac12\bar{\alpha}_TM^2
        }{1-\bar{\alpha}_T}
        =
        \kappa_T.
    \end{aligned}
\end{equation}
Consequently,
\begin{equation}
\label{eq:initial_unconditional_weight_bounds}
    e^{-\kappa_T}\le w_T(\mathbf{X})\le e^{\kappa_T},
    \qquad
    e^{-\kappa_T}\le Z_T\le e^{\kappa_T},
\end{equation}
and hence, $q$-almost surely,
\begin{equation}
\label{eq:initial_unconditional_normalized_weight}
    \begin{aligned}
        e^{-2\kappa_T}
        &\le
        \frac{w_T(\mathbf{X})}{Z_T}
        \le
        e^{2\kappa_T},
        \\
        \left|\frac{w_T(\mathbf{X})}{Z_T}-1\right|
        &\le
        \max\!\left\{
            1-e^{-2\kappa_T},e^{2\kappa_T}-1
        \right\}
        =
        e^{2\kappa_T}-1.
    \end{aligned}
\end{equation}
Subtracting $\boldsymbol{\mu}=\mathbb{E}_{\mathbf{X}\sim q}[\mathbf{X}]$ from \autoref{eq:initial_unconditional_weighted_mean} gives
\begin{equation}
\label{eq:initial_unconditional_mean_difference}
    \hat{\mathbf{x}}_{T,\emptyset}-\boldsymbol{\mu}
    =
    \mathbb{E}_{\mathbf{X}\sim q}\!\left[
        \mathbf{X}\left(\frac{w_T(\mathbf{X})}{Z_T}-1\right)
    \right].
\end{equation}
Therefore,
\begin{equation}
\label{eq:initial_unconditional_error_bound}
    \begin{aligned}
        \left\|\hat{\mathbf{x}}_{T,\emptyset}-\boldsymbol{\mu}\right\|_2
        &\le
        \mathbb{E}_{\mathbf{X}\sim q}\!\left[
            \|\mathbf{X}\|_2
            \left|\frac{w_T(\mathbf{X})}{Z_T}-1\right|
        \right]
        \\
        &\le
        M\left(e^{2\kappa_T}-1\right).
    \end{aligned}
\end{equation}
At fixed $\mathbf{x}_T$, we have $\kappa_T\to0$ as $\bar{\alpha}_T\to0$, so the bound tends to zero.
\end{proof}
\subsection{Proof of \autoref{thm:initial_cfg_amplification}}
\label{app:initial_cfg_amplification}

\begin{theorem}[Initial CFG amplification]
Assume the conditions of \autoref{lem:initial_unconditional_prediction} and $\mathbb{E}_{\boldsymbol{\epsilon}}[\mathcal{L}_c(\mathbf{x}^\star,T)]\le\varepsilon_T$.
For $\mathbf{x}_T\sim\mathcal{N}(\mathbf{0},\mathbf{I})$, $g>1$, and any tolerance $r>0$,
\[
    \Pr\!\left(
        \left\|
            \hat{\mathbf{x}}_{T,g}
            -\left[\boldsymbol{\mu}+g(\mathbf{x}^\star-\boldsymbol{\mu})\right]
        \right\|_2>r
    \right)
    \le
    g^2\frac{1-\bar{\alpha}_T}{\bar{\alpha}_T}
    \frac{\varepsilon_T}{r^2}
    +o(1),
\]
as $\bar{\alpha}_T\to0$.
\end{theorem}

\begin{proof}
For each realized initial latent $\mathbf{x}_T$, the CFG decomposition in \autoref{eq:predx} gives
\begin{equation}
\label{eq:initial_cfg_error_decomposition}
    \begin{aligned}
        &\hat{\mathbf{x}}_{T,g}
        -\left[\boldsymbol{\mu}+g(\mathbf{x}^\star-\boldsymbol{\mu})\right]
        \\
        &\qquad=
        g\left(\hat{\mathbf{x}}_{T,c}-\mathbf{x}^\star\right)
        +(1-g)\left(\hat{\mathbf{x}}_{T,\emptyset}-\boldsymbol{\mu}\right).
    \end{aligned}
\end{equation}
Since $g>1$, the triangle inequality and \autoref{lem:initial_unconditional_prediction} imply
\begin{equation}
\label{eq:initial_cfg_pointwise_bound}
    \begin{aligned}
        &\left\|
            \hat{\mathbf{x}}_{T,g}
            -\left[\boldsymbol{\mu}+g(\mathbf{x}^\star-\boldsymbol{\mu})\right]
        \right\|_2
        \\
        &\qquad\le
        g\left\|\hat{\mathbf{x}}_{T,c}-\mathbf{x}^\star\right\|_2
        +(g-1)\left\|\hat{\mathbf{x}}_{T,\emptyset}-\boldsymbol{\mu}\right\|_2
        \\
        &\qquad\le
        g\left\|\hat{\mathbf{x}}_{T,c}-\mathbf{x}^\star\right\|_2
        +(g-1)M\left(e^{2\kappa_T}-1\right).
    \end{aligned}
\end{equation}

By \autoref{eq:branch_posterior_mean, eq:bounded_posterior}, $\|\hat{\mathbf{x}}_{T,b}(\mathbf{z})\|_2\le M$ for $b\in\{c,\emptyset\}$.
Thus, the conditional squared error satisfies
\begin{equation}
\label{eq:initial_cfg_conditional_error_bound}
    0
    \le
    \left\|\hat{\mathbf{x}}_{T,c}(\mathbf{z})-\mathbf{x}^\star\right\|_2^2
    \le
    \left(M+\|\mathbf{x}^\star\|_2\right)^2
\end{equation}
at every input $\mathbf{z}$.
For a nonnegative bounded function, the absolute difference between its expectations under two distributions is at most its upper bound times their total variation distance.
Applying \autoref{eq:initial_conditional_reproduction_mse, eq:initial_conditional_reproduction_tv} therefore gives
\begin{equation}
\label{eq:initial_cfg_conditional_second_moment}
    \begin{aligned}
        &\mathbb{E}_{\mathbf{x}_T}\!\left[
            \left\|\hat{\mathbf{x}}_{T,c}-\mathbf{x}^\star\right\|_2^2
        \right]
        \\
        &\qquad\le
        \frac{1-\bar{\alpha}_T}{\bar{\alpha}_T}\varepsilon_T
        \\
        &\qquad\quad+
        \frac{\left(M+\|\mathbf{x}^\star\|_2\right)^2}{2}
        \sqrt{
            \bar{\alpha}_T\|\mathbf{x}^\star\|_2^2
            +d\left[-\bar{\alpha}_T-\log(1-\bar{\alpha}_T)\right]
        }
        \\
        &\qquad=
        \frac{1-\bar{\alpha}_T}{\bar{\alpha}_T}\varepsilon_T
        +o(1).
    \end{aligned}
\end{equation}

For the unconditional branch, $\|\boldsymbol{\mu}\|_2\le\mathbb{E}_{\mathbf{X}\sim q}[\|\mathbf{X}\|_2]\le M$ and \autoref{lem:initial_unconditional_prediction} yield
\begin{equation}
\label{eq:initial_cfg_unconditional_error_bound}
    \left\|\hat{\mathbf{x}}_{T,\emptyset}-\boldsymbol{\mu}\right\|_2
    \le
    \min\!\left\{
        2M,\,
        M\left(e^{2\kappa_T}-1\right)
    \right\}
    \longrightarrow 0
\end{equation}
as $\bar{\alpha}_T\to0$ at each fixed $\mathbf{x}_T$.
Since the initialization distribution remains $\mathcal{N}(\mathbf{0},\mathbf{I})$, dominated convergence gives
\begin{equation}
\label{eq:initial_cfg_unconditional_moments}
    \begin{aligned}
        \mathbb{E}_{\mathbf{x}_T}\!\left[
            \left\|\hat{\mathbf{x}}_{T,\emptyset}-\boldsymbol{\mu}\right\|_2
        \right]
        &\longrightarrow 0,
        \\
        \mathbb{E}_{\mathbf{x}_T}\!\left[
            \left\|\hat{\mathbf{x}}_{T,\emptyset}-\boldsymbol{\mu}\right\|_2^2
        \right]
        &\longrightarrow 0.
    \end{aligned}
\end{equation}

Squaring the first inequality in \autoref{eq:initial_cfg_pointwise_bound}, taking expectation, and using these bounds yields
\begin{equation}
\label{eq:initial_cfg_second_moment}
    \begin{aligned}
        &\mathbb{E}_{\mathbf{x}_T}\!\left[
            \left\|
                \hat{\mathbf{x}}_{T,g}
                -\left[\boldsymbol{\mu}+g(\mathbf{x}^\star-\boldsymbol{\mu})\right]
            \right\|_2^2
        \right]
        \\
        &\qquad\le
        g^2\mathbb{E}_{\mathbf{x}_T}\!\left[
            \left\|\hat{\mathbf{x}}_{T,c}-\mathbf{x}^\star\right\|_2^2
        \right]
        \\
        &\qquad\quad+
        2g(g-1)\left(M+\|\mathbf{x}^\star\|_2\right)
        \mathbb{E}_{\mathbf{x}_T}\!\left[
            \left\|\hat{\mathbf{x}}_{T,\emptyset}-\boldsymbol{\mu}\right\|_2
        \right]
        \\
        &\qquad\quad+
        (g-1)^2\mathbb{E}_{\mathbf{x}_T}\!\left[
            \left\|\hat{\mathbf{x}}_{T,\emptyset}-\boldsymbol{\mu}\right\|_2^2
        \right]
        \\
        &\qquad\le
        g^2\frac{1-\bar{\alpha}_T}{\bar{\alpha}_T}\varepsilon_T
        +o(1).
    \end{aligned}
\end{equation}
Finally, Markov's inequality gives
\begin{equation}
\label{eq:initial_cfg_probability_bound}
    \begin{aligned}
        &\Pr_{\mathbf{x}_T}\!\left(
            \left\|
                \hat{\mathbf{x}}_{T,g}
                -\left[\boldsymbol{\mu}+g(\mathbf{x}^\star-\boldsymbol{\mu})\right]
            \right\|_2>r
        \right)
        \\
        &\qquad\le
        \frac{1}{r^2}
        \mathbb{E}_{\mathbf{x}_T}\!\left[
            \left\|
                \hat{\mathbf{x}}_{T,g}
                -\left[\boldsymbol{\mu}+g(\mathbf{x}^\star-\boldsymbol{\mu})\right]
            \right\|_2^2
        \right]
        \\
        &\qquad\le
        g^2\frac{1-\bar{\alpha}_T}{\bar{\alpha}_T}
        \frac{\varepsilon_T}{r^2}
        +o(1),
    \end{aligned}
\end{equation}
as $\bar{\alpha}_T\to0$ for each fixed tolerance $r>0$.
\end{proof}

\textbf{Finite-model correction.}
At the first denoising step, \autoref{eq:initial_cfg_error_decomposition} holds without assuming \autoref{eq:bayes_optimal}.
Under $\|\mathbf{X}\|_2\le M$ $q$-almost surely, adding and subtracting the unconditional posterior mean and applying its bound from \autoref{app:initial_unconditional_prediction} gives
\begin{equation}
\label{eq:initial_cfg_finite_model_bound}
    \begin{aligned}
        &\left\|
            \hat{\mathbf{x}}_{T,g}
            -\left[\boldsymbol{\mu}
                +g(\mathbf{x}^\star-\boldsymbol{\mu})\right]
        \right\|_2
        \\
        &\qquad\le
        g\left\|\hat{\mathbf{x}}_{T,c}-\mathbf{x}^\star\right\|_2
        +(g-1)M\left(e^{2\kappa_T}-1\right)
        \\
        &\qquad\quad+
        (g-1)\left\|
            \hat{\mathbf{x}}_{T,\emptyset}
            -\mathbb{E}_{\mathbf{X}}\!\left[
                \mathbf{X}
                \mid\mathbf{X}_T=\mathbf{x}_T,T,\emptyset
            \right]
        \right\|_2.
    \end{aligned}
\end{equation}
The additional term measures the unconditional clean prediction's departure from its posterior mean.
Conditional reproduction is already quantified through the actual network loss in \autoref{lem:initial_conditional_reproduction}, so no separate conditional model-error term is needed.
\subsection{Proof of \autoref{prop:subsequent_conditional_reproduction}}
\label{app:subsequent_conditional_reproduction}

\begin{proposition}[Subsequent conditional reproduction]
Fix $t\in\{1,\ldots,T\}$, a realized sampling latent $\mathbf{x}_t$, and a tolerance $r\ge0$.
Then $\|\hat{\mathbf{x}}_{t,c}-\mathbf{x}^\star\|_2\le r$ if and only if
\[
    \left\|
        \frac{\mathbf{x}_t-\sqrt{\bar{\alpha}_t}\mathbf{x}^\star}
             {\sqrt{1-\bar{\alpha}_t}}
        -\boldsymbol{\epsilon}_c(\mathbf{x}_t,t)
    \right\|_2
    \le r\sqrt{\frac{\bar{\alpha}_t}{1-\bar{\alpha}_t}}.
\]
\end{proposition}

\begin{proof}
Using \autoref{eq:branch_clean_prediction} at $\mathbf{x}_t$ and subtracting $\mathbf{x}^\star$ gives
\begin{equation}
\label{eq:subsequent_conditional_reproduction_error_identity}
    \begin{aligned}
        \hat{\mathbf{x}}_{t,c}-\mathbf{x}^\star
        &=
        \frac{
            \mathbf{x}_t
            -\sqrt{1-\bar{\alpha}_t}\,\boldsymbol{\epsilon}_c(\mathbf{x}_t,t)
            -\sqrt{\bar{\alpha}_t}\mathbf{x}^\star
        }{\sqrt{\bar{\alpha}_t}}
        \\
        &=
        \sqrt{\frac{1-\bar{\alpha}_t}{\bar{\alpha}_t}}
        \left[
            \frac{\mathbf{x}_t-\sqrt{\bar{\alpha}_t}\mathbf{x}^\star}
                 {\sqrt{1-\bar{\alpha}_t}}
            -\boldsymbol{\epsilon}_c(\mathbf{x}_t,t)
        \right].
    \end{aligned}
\end{equation}
Since $0<\bar{\alpha}_t<1$, taking Euclidean norms and rearranging proves the equivalence.
\end{proof}
\subsection{Proof of \autoref{thm:propagation_of_prompt_influence}}
\label{app:propagation_of_prompt_influence}

\begin{theorem}[Propagation of prompt influence]
For $t\in\{2,\ldots,T\}$, suppose $\hat{\mathbf{x}}_{t-1,\emptyset}(\cdot)$ is continuously differentiable on a neighborhood of the segment in \autoref{eq:propagation_segment}.
Then
\[
    \hat{\mathbf{x}}_{t-1,\emptyset}(\mathbf{x}_{t-1})
    -\hat{\mathbf{x}}_{t-1,\emptyset}(\mathbf{x}_{t-1}^{\emptyset})
    =g\eta_t\int_0^1
    \mathbf{J}_{t-1,\emptyset}(\mathbf{z}_t(s))
    \boldsymbol{\Delta}_t\,ds,
\]
and consequently
\[
    \left\|
        \hat{\mathbf{x}}_{t-1,\emptyset}(\mathbf{x}_{t-1})
        -\hat{\mathbf{x}}_{t-1,\emptyset}(\mathbf{x}_{t-1}^{\emptyset})
    \right\|_2
    \le g\eta_t\sup_{s\in[0,1]}
    \left\|\mathbf{J}_{t-1,\emptyset}(\mathbf{z}_t(s))\right\|_{\mathrm{op}}
    \|\boldsymbol{\Delta}_t\|_2.
\]
\end{theorem}

\begin{proof}
The segment in \autoref{eq:propagation_segment} satisfies $\mathbf{z}_t'(s)=g\eta_t\boldsymbol{\Delta}_t$, so the chain rule gives
\begin{equation}
\label{eq:propagation_chain_rule}
    \frac{d}{ds}\hat{\mathbf{x}}_{t-1,\emptyset}(\mathbf{z}_t(s))
    =\mathbf{J}_{t-1,\emptyset}(\mathbf{z}_t(s))\mathbf{z}_t'(s)
    =g\eta_t\mathbf{J}_{t-1,\emptyset}(\mathbf{z}_t(s))\boldsymbol{\Delta}_t.
\end{equation}
Integrating over $s\in[0,1]$ and using $\mathbf{z}_t(0)=\mathbf{x}_{t-1}^{\emptyset}$ and $\mathbf{z}_t(1)=\mathbf{x}_{t-1}$ proves the stated identity by the fundamental theorem of calculus.
Since $g\eta_t>0$, the triangle and operator-norm inequalities give
\begin{equation}
\label{eq:propagation_prediction_change_bound}
    \begin{aligned}
        &\left\|
            \hat{\mathbf{x}}_{t-1,\emptyset}(\mathbf{x}_{t-1})
            -\hat{\mathbf{x}}_{t-1,\emptyset}(\mathbf{x}_{t-1}^{\emptyset})
        \right\|_2
        \\
        &\qquad\le
        g\eta_t\int_0^1
        \left\|
            \mathbf{J}_{t-1,\emptyset}(\mathbf{z}_t(s))
            \boldsymbol{\Delta}_t
        \right\|_2\,ds
        \\
        &\qquad\le
        g\eta_t\int_0^1
        \left\|\mathbf{J}_{t-1,\emptyset}(\mathbf{z}_t(s))\right\|_{\mathrm{op}}
        \|\boldsymbol{\Delta}_t\|_2\,ds
        \\
        &\qquad\le
        g\eta_t\sup_{s\in[0,1]}
        \left\|\mathbf{J}_{t-1,\emptyset}(\mathbf{z}_t(s))\right\|_{\mathrm{op}}
        \|\boldsymbol{\Delta}_t\|_2.
    \end{aligned}
\end{equation}
The supremum is finite because the Jacobian is continuous on the compact segment.
\end{proof}

\textbf{Continuous differentiability of $\hat{\mathbf{x}}_{t-1,\emptyset}(\cdot)$.}
Under \autoref{eq:bayes_optimal}, the unconditional clean prediction equals the posterior mean in \autoref{eq:branch_posterior_mean}.
For $t\ge2$, the Gaussian forward likelihood at timestep $t-1$ has positive variance $1-\bar{\alpha}_{t-1}>0$.
Bayes' formula expresses this posterior mean as a ratio of smooth Gaussian integrals with a strictly positive denominator, ensuring continuous differentiability.
\subsection{Proof of \autoref{thm:final_reproduction}}
\label{app:final_reproduction}

\begin{theorem}[Final reproduction]
Assume \autoref{eq:bayes_optimal} and $\|\mathbf{X}\|_2\le M$ almost surely under the conditional and unconditional data distributions.
For each realized latent $\mathbf{x}_1$ at the final denoising step and $g>1$, the final generated latent satisfies
\[
    \|\mathbf{x}_0-\mathbf{x}^\star\|_2
    \le\bigl(M+\|\mathbf{x}^\star\|_2\bigr)
    \Bigl[
        g\bigl(1-\pi_{1,c}(\mathbf{x}_1)\bigr)
        +(g-1)\bigl(1-\pi_{1,\emptyset}(\mathbf{x}_1)\bigr)
    \Bigr].
\]
\end{theorem}

\begin{proof}
Fix the realized latent $\mathbf{x}_1$ at the final denoising step.
Under \autoref{eq:bayes_optimal}, each clean prediction equals the posterior mean in \autoref{eq:branch_posterior_mean} evaluated at $\mathbf{x}_1$.
For $b\in\{c,\emptyset\}$ with $\pi_{1,b}(\mathbf{x}_1)<1$, subtracting $\mathbf{x}^\star$ from \autoref{eq:final_posterior_decomposition} and taking norms gives
\begin{equation}
\label{eq:final_branch_reproduction_error}
    \|\hat{\mathbf{x}}_{1,b}-\mathbf{x}^\star\|_2
    =\bigl[1-\pi_{1,b}(\mathbf{x}_1)\bigr]
    \left\|
        \mathbb{E}_{\mathbf{X}}\!\left[
            \mathbf{X}
            \mid\mathbf{X}_1=\mathbf{x}_1,1,b,
            \mathbf{X}\ne\mathbf{x}^\star
        \right]-\mathbf{x}^\star
    \right\|_2.
\end{equation}
If $\pi_{1,b}(\mathbf{x}_1)=1$, then $\hat{\mathbf{x}}_{1,b}=\mathbf{x}^\star$, and the weighted non-target term is defined as zero.

The identity $\mathbf{x}_0=\hat{\mathbf{x}}_{1,g}$ and \autoref{eq:predx} give
\begin{equation}
\label{eq:final_cfg_error_decomposition}
    \mathbf{x}_0-\mathbf{x}^\star
    =g\bigl(\hat{\mathbf{x}}_{1,c}-\mathbf{x}^\star\bigr)
    +(1-g)\bigl(\hat{\mathbf{x}}_{1,\emptyset}-\mathbf{x}^\star\bigr).
\end{equation}
Since $g>1$, the triangle inequality and \autoref{eq:final_branch_reproduction_error} imply
\begin{equation}
\label{eq:final_general_reproduction_bound}
    \begin{aligned}
        \|\mathbf{x}_0-\mathbf{x}^\star\|_2
        &\le
        g\|\hat{\mathbf{x}}_{1,c}-\mathbf{x}^\star\|_2
        +(g-1)\|\hat{\mathbf{x}}_{1,\emptyset}-\mathbf{x}^\star\|_2
        \\
        &=g\bigl[1-\pi_{1,c}(\mathbf{x}_1)\bigr]
        \left\|
            \mathbb{E}_{\mathbf{X}}\!\left[
                \mathbf{X}
                \mid\mathbf{X}_1=\mathbf{x}_1,1,c,
                \mathbf{X}\ne\mathbf{x}^\star
            \right]-\mathbf{x}^\star
        \right\|_2
        \\
        &\quad+(g-1)\bigl[1-\pi_{1,\emptyset}(\mathbf{x}_1)\bigr]
        \left\|
            \mathbb{E}_{\mathbf{X}}\!\left[
                \mathbf{X}
                \mid\mathbf{X}_1=\mathbf{x}_1,1,\emptyset,
                \mathbf{X}\ne\mathbf{x}^\star
            \right]-\mathbf{x}^\star
        \right\|_2.
    \end{aligned}
\end{equation}
This bound retains the full non-target contribution and does not use the assumption $\|\mathbf{X}\|_2\le M$.

Since $0<\bar{\alpha}_1<1$, the Gaussian forward likelihood is strictly positive, and Bayes' rule preserves each prior's almost-sure bound $\|\mathbf{X}\|_2\le M$ under posterior conditioning.
For $\pi_{1,b}(\mathbf{x}_1)<1$, the triangle inequality and Jensen's inequality therefore give
\begin{equation}
\label{eq:final_non_target_displacement_bound}
    \begin{aligned}
        &\left\|
            \mathbb{E}_{\mathbf{X}}\!\left[
                \mathbf{X}
                \mid\mathbf{X}_1=\mathbf{x}_1,1,b,
                \mathbf{X}\ne\mathbf{x}^\star
            \right]-\mathbf{x}^\star
        \right\|_2
        \\
        &\qquad\le
        \mathbb{E}_{\mathbf{X}}\!\left[
            \|\mathbf{X}\|_2
            \mid\mathbf{X}_1=\mathbf{x}_1,1,b,
            \mathbf{X}\ne\mathbf{x}^\star
        \right]+\|\mathbf{x}^\star\|_2
        \\
        &\qquad\le M+\|\mathbf{x}^\star\|_2.
    \end{aligned}
\end{equation}
Substituting \autoref{eq:final_non_target_displacement_bound} into \autoref{eq:final_general_reproduction_bound} proves the theorem.
\end{proof}
\subsection{Proof of \autoref{lem:posterior_probability_gap}}
\label{app:posterior_probability_gap}

\begin{lemma}[Posterior probability gap]
Assume $0<p_{\emptyset}<p_c<1$.
The posterior probability gap $G(\lambda)$ is unimodal in the target evidence $\lambda$, with unique maximizer
\[
    \lambda^\star
    :=-\frac{\operatorname{logit}(p_c)+\operatorname{logit}(p_{\emptyset})}{2}.
\]
On $[\lambda_T,\infty)$, $G$ strictly decreases if $\lambda_T\ge\lambda^\star$.
Otherwise, it strictly increases on $[\lambda_T,\lambda^\star]$ and strictly decreases on $[\lambda^\star,\infty)$.
When $\lambda_T=0$, these cases correspond to $p_c+p_{\emptyset}\ge1$ and $p_c+p_{\emptyset}<1$, respectively.
\end{lemma}

\begin{proof}
Differentiating \autoref{eq:posterior_probability_gap} gives
\begin{equation}
\label{eq:posterior_probability_gap_derivative}
    G'(\lambda)
    =\sigma'\!\left(\lambda+\operatorname{logit}(p_c)\right)
    -\sigma'\!\left(\lambda+\operatorname{logit}(p_{\emptyset})\right).
\end{equation}
The sigmoid derivative satisfies
\begin{equation}
\label{eq:posterior_probability_gap_sigmoid_derivative}
    \sigma'(u)
    =\sigma(u)\bigl[1-\sigma(u)\bigr]
    =\frac{1}{4\cosh^2(u/2)},
\end{equation}
so $\sigma'$ is even and strictly decreases as $|u|$ increases.
Since $\operatorname{logit}(p_c)>\operatorname{logit}(p_{\emptyset})$, comparing the squared arguments yields
\begin{equation}
\label{eq:posterior_probability_gap_derivative_sign}
    \begin{aligned}
        G'(\lambda)>0
        &\quad\Longleftrightarrow\quad
        \left|\lambda+\operatorname{logit}(p_c)\right|
        <\left|\lambda+\operatorname{logit}(p_{\emptyset})\right|
        \\
        &\quad\Longleftrightarrow\quad
        2\lambda+\operatorname{logit}(p_c)
        +\operatorname{logit}(p_{\emptyset})<0
        \\
        &\quad\Longleftrightarrow\quad
        \lambda<\lambda^\star.
    \end{aligned}
\end{equation}
Similarly, $G'(\lambda)<0$ for $\lambda>\lambda^\star$, and $G'(\lambda^\star)=0$.
Thus, $G$ strictly increases on $(-\infty,\lambda^\star]$ and strictly decreases on $[\lambda^\star,\infty)$, proving unimodality and the unique maximizer.

By the definition of $\lambda^\star$,
\begin{equation}
\label{eq:posterior_probability_gap_initial_evidence}
    \lambda_T\ge\lambda^\star
    \quad\Longleftrightarrow\quad
    2\lambda_T+\operatorname{logit}(p_c)
    +\operatorname{logit}(p_{\emptyset})\ge0.
\end{equation}
If $\lambda_T\ge\lambda^\star$, then $[\lambda_T,\infty)$ lies within $[\lambda^\star,\infty)$, where $G$ is strictly decreasing.
Otherwise, $\lambda_T<\lambda^\star$, so $G$ strictly increases on $[\lambda_T,\lambda^\star]$ and strictly decreases on $[\lambda^\star,\infty)$.

When $\lambda_T=0$, the condition $\lambda_T\ge\lambda^\star$ becomes
\begin{equation}
\label{eq:posterior_probability_gap_zero_evidence}
    \begin{aligned}
        \operatorname{logit}(p_c)+\operatorname{logit}(p_{\emptyset})\ge0
        &\quad\Longleftrightarrow\quad
        \log\frac{p_cp_{\emptyset}}{(1-p_c)(1-p_{\emptyset})}\ge0
        \\
        &\quad\Longleftrightarrow\quad
        p_cp_{\emptyset}\ge(1-p_c)(1-p_{\emptyset})
        \\
        &\quad\Longleftrightarrow\quad
        p_c+p_{\emptyset}\ge1.
    \end{aligned}
\end{equation}
The complementary case $\lambda_T<\lambda^\star$ therefore corresponds to $p_c+p_{\emptyset}<1$.
\end{proof}
\subsection{Proof of \autoref{thm:posterior_lag_dynamics}}
\label{app:posterior_lag_dynamics}

\begin{theorem}[Posterior lag dynamics]
Assume \autoref{eq:bayes_optimal} and $0<p_{\emptyset}<p_c<1$.
For each timestep $t\in\{1,\ldots,T\}$ and realized latent $\mathbf{x}_t$ satisfying \autoref{eq:non_target_likelihood_equality}, the posterior lag and its magnitude satisfy
\[
    \begin{aligned}
        \left|
            \|\boldsymbol{\Delta}_t\|_2
            -G(\lambda_t)\|\mathbf{x}^\star\|_2
        \right|
        &\le
        \|\boldsymbol{\Delta}_t-G(\lambda_t)\mathbf{x}^\star\|_2
        \\
        &\le
        \sum_{b\in\{c,\emptyset\}}
        \bigl[1-\pi_{t,b}(\mathbf{x}_t)\bigr]
        \left\|
            \mathbb{E}_{\mathbf{X}}\!\left[
                \mathbf{X}
                \mid\mathbf{X}_t=\mathbf{x}_t,t,b,
                \mathbf{X}\ne\mathbf{x}^\star
            \right]
        \right\|_2.
    \end{aligned}
\]
\end{theorem}

\begin{proof}
Fix $t\in\{1,\ldots,T\}$ and a realized sampling latent $\mathbf{x}_t$ satisfying \autoref{eq:non_target_likelihood_equality}.
Since $0<\bar{\alpha}_t<1$, the Gaussian forward likelihood is strictly positive, and $0<p_b<1$ ensures $0<\pi_{t,b}(\mathbf{x}_t)<1$ for $b\in\{c,\emptyset\}$.
Bayes' rule, \autoref{eq:non_target_likelihood_equality}, and the definition of $\lambda_t$ give
\begin{equation}
\label{eq:posterior_lag_posterior_odds}
    \begin{aligned}
        \frac{\pi_{t,b}(\mathbf{x}_t)}{1-\pi_{t,b}(\mathbf{x}_t)}
        &=\frac{p_b}{1-p_b}
        \frac{
            q(\mathbf{x}_t\mid\mathbf{x}^\star)
        }{
            \mathbb{E}_{\mathbf{X}}\!\left[
                q(\mathbf{x}_t\mid\mathbf{X})
                \mid b,\mathbf{X}\ne\mathbf{x}^\star
            \right]
        }
        \\
        &=\exp\!\left(\lambda_t+\operatorname{logit}(p_b)\right).
    \end{aligned}
\end{equation}
Taking logarithms and applying $\sigma$ gives \autoref{eq:target_posterior_from_evidence}.
Since $p_c>p_{\emptyset}$ and $\sigma$ is strictly increasing, \autoref{eq:posterior_probability_gap} yields $\pi_{t,c}(\mathbf{x}_t)-\pi_{t,\emptyset}(\mathbf{x}_t)=G(\lambda_t)>0$.

Under \autoref{eq:bayes_optimal}, each clean prediction equals the forward-model posterior mean in \autoref{eq:branch_posterior_mean} evaluated at $\mathbf{x}_t$.
For $b\in\{c,\emptyset\}$, the law of total expectation gives
\begin{equation}
\label{eq:posterior_lag_branch_decomposition}
    \begin{aligned}
        \hat{\mathbf{x}}_{t,b}
        &=\pi_{t,b}(\mathbf{x}_t)\mathbf{x}^\star
        \\
        &\quad+\bigl[1-\pi_{t,b}(\mathbf{x}_t)\bigr]
        \mathbb{E}_{\mathbf{X}}\!\left[
            \mathbf{X}
            \mid\mathbf{X}_t=\mathbf{x}_t,t,b,
            \mathbf{X}\ne\mathbf{x}^\star
        \right].
    \end{aligned}
\end{equation}
Subtracting the unconditional decomposition from the conditional decomposition and using $\boldsymbol{\Delta}_t=\hat{\mathbf{x}}_{t,c}-\hat{\mathbf{x}}_{t,\emptyset}$ yields
\begin{equation}
\label{eq:posterior_lag_non_target_difference}
    \begin{aligned}
        \boldsymbol{\Delta}_t-G(\lambda_t)\mathbf{x}^\star
        &=\bigl[1-\pi_{t,c}(\mathbf{x}_t)\bigr]
        \mathbb{E}_{\mathbf{X}}\!\left[
            \mathbf{X}
            \mid\mathbf{X}_t=\mathbf{x}_t,t,c,
            \mathbf{X}\ne\mathbf{x}^\star
        \right]
        \\
        &\quad-\bigl[1-\pi_{t,\emptyset}(\mathbf{x}_t)\bigr]
        \mathbb{E}_{\mathbf{X}}\!\left[
            \mathbf{X}
            \mid\mathbf{X}_t=\mathbf{x}_t,t,\emptyset,
            \mathbf{X}\ne\mathbf{x}^\star
        \right].
    \end{aligned}
\end{equation}
Since $G(\lambda_t)>0$, the reverse triangle inequality and the triangle inequality applied to \autoref{eq:posterior_lag_non_target_difference} give
\begin{equation}
\label{eq:posterior_lag_magnitude_bound}
    \begin{aligned}
        \left|
            \|\boldsymbol{\Delta}_t\|_2
            -G(\lambda_t)\|\mathbf{x}^\star\|_2
        \right|
        &\le
        \|\boldsymbol{\Delta}_t-G(\lambda_t)\mathbf{x}^\star\|_2
        \\
        &\le
        \sum_{b\in\{c,\emptyset\}}
        \bigl[1-\pi_{t,b}(\mathbf{x}_t)\bigr]
        \left\|
            \mathbb{E}_{\mathbf{X}}\!\left[
                \mathbf{X}
                \mid\mathbf{X}_t=\mathbf{x}_t,t,b,
                \mathbf{X}\ne\mathbf{x}^\star
            \right]
        \right\|_2.
    \end{aligned}
\end{equation}
This proves both inequalities.
\end{proof}
\section{Dataset}
\label{app:dataset}

In this paper, we use prompts from \citet{webster2023extraction}, sourced from the LAION-5B dataset~\citep{schuhmann2022laion5b}, for Stable Diffusion v1.4 (SDv1)~\citep{rombach2022sdv1}, Stable Diffusion v2.1 (SDv2)~\citep{lopez2022sdv2}, and RealisticVision (RealVis)~\citep{sg1612222023realvis}.
The released metadata group the prompts into four categories:

\textbf{(i) Matching verbatim (\texttt{MV})}: generated outputs that closely reproduce their paired training images;

\textbf{(ii) Retrieval verbatim (\texttt{RV})}: cases where the output does not match the paired training image but reproduces another training image identified through retrieval;

\textbf{(iii) Template verbatim (\texttt{TV})}: images that share the overall template with a training image but differ in details such as colors or textures; and

\textbf{(iv) None (\texttt{N})}: cases where no verbatim reproduction was identified in the original evaluation.

Each prompt was provided with a URL linking to its paired training image.
However, some URLs were inaccessible, preventing direct retrieval of the corresponding images.
For these cases, our data preparation procedure also checks a ground-truth mirror, accepting an image only when its original index, prompt, and URL match the source entry.\footnote{\url{https://huggingface.co/datasets/gdhanuka/memorization_data2}}
We exclude pairs whose target images remain unavailable and use all remaining pairs, as shown in \autoref{tab:dataset}.
Each prompt's paired target image remains fixed throughout the experiments.

The original categories describe generation outcomes under the settings and evaluation criteria of \citet{webster2023extraction}, which can differ from those used in our experiments.
We therefore re-score prompts using SSCD~\citep{pizzi2022sscd} to assess reproduction of each prompt's paired target under our model and sampler configurations (\autoref{sec:settings}).
This provides a consistent criterion for grouping prompts across all configurations.
For each model--sampler configuration, we classify a prompt as memorized if at least one of its 20 generated images has SSCD $>0.7$ with its paired target, and as normal if all scores are $\leq0.7$.
These groups are used for the experimental comparisons, while the categories in \autoref{tab:dataset} retain the original source annotations.
All generated samples remain in their prompt's group, and prompt-group summaries give each prompt equal weight.

\begin{table}[ht]
    \centering
    \small
    \caption{
        \textbf{Dataset.}
        Original prompt categories and counts across different diffusion models.
        Fractions indicate the number of prompts with available paired target images over the total number of prompts originally provided by \citet{webster2023extraction}.
    }
    \label{tab:dataset}
    \vspace{2.0mm}
    \begin{tabular}{rccccc}
        \toprule
        & \texttt{MV} & \texttt{RV} & \texttt{TV} & \texttt{N} & Total \\
        \midrule
        SDv1 & 86/86 & 30/30 & 229/229 & 101/155 & 446/500 \\
        SDv2 & 2/4 & 0/0 & 208/215 & 180/281 & 390/500 \\
        RealVis & 89/90 & 34/34 & 229/230 & 94/146 & 446/500 \\
        \bottomrule
    \end{tabular}
\end{table}
\section{Additional results}
\label{app:additional_results}

\begin{figure}[t]
    \centering
    \captionsetup[subfigure]{skip=2pt,hypcap=true}
    \setlength{\parskip}{0pt}

    \begin{subfigure}[t]{0.32\linewidth}
        \centering
        \includegraphics[
            width=\linewidth,
            viewport=-6 -7 351.78 327.464,
            clip
        ]{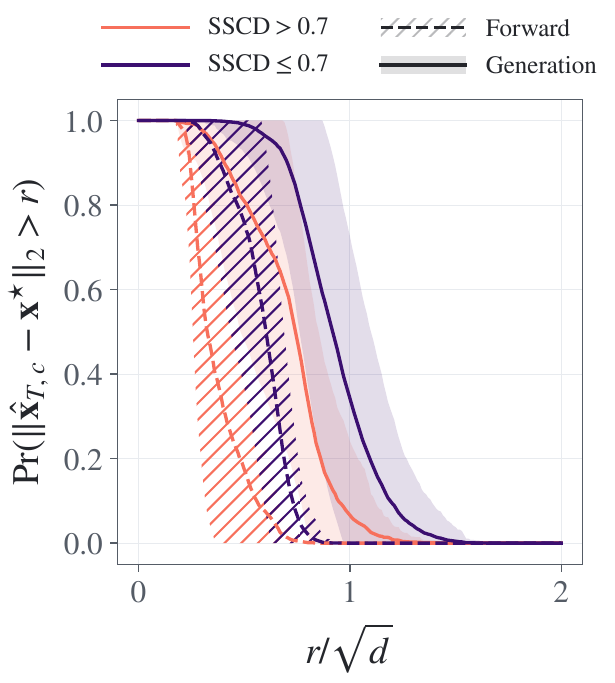}
        \caption{\autoref{lem:initial_conditional_reproduction}}
        \label{fig:sdv1_ddpm_lem1}
    \end{subfigure}\hfill
    \begin{subfigure}[t]{0.32\linewidth}
        \centering
        \includegraphics[
            width=\linewidth,
            viewport=-2 -10.25 355.78 324.214,
            clip
        ]{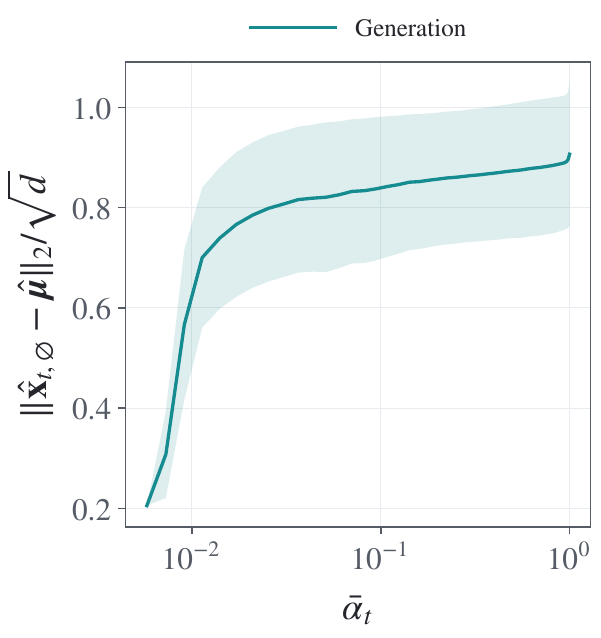}
        \caption{\autoref{lem:initial_unconditional_prediction}}
        \label{fig:sdv1_ddpm_lem2}
    \end{subfigure}\hfill
    \begin{subfigure}[t]{0.32\linewidth}
        \centering
        \includegraphics[
            width=\linewidth,
            viewport=-5 -7 352.78 327.464,
            clip
        ]{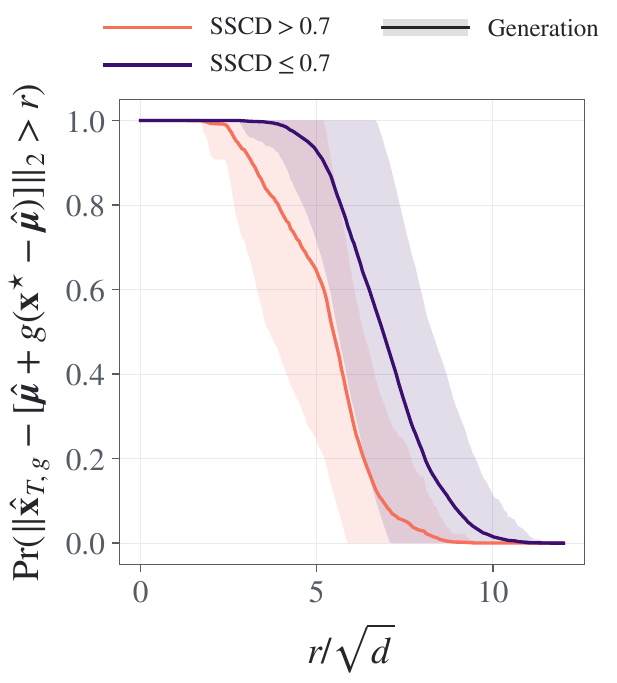}
        \caption{\autoref{thm:initial_cfg_amplification}}
        \label{fig:sdv1_ddpm_thm3}
    \end{subfigure}

    \par\vspace{6pt}

    \begin{subfigure}[t]{0.32\linewidth}
        \centering
        \includegraphics[
            width=\linewidth,
            viewport=-2 -13 355.78 321.464,
            clip
        ]{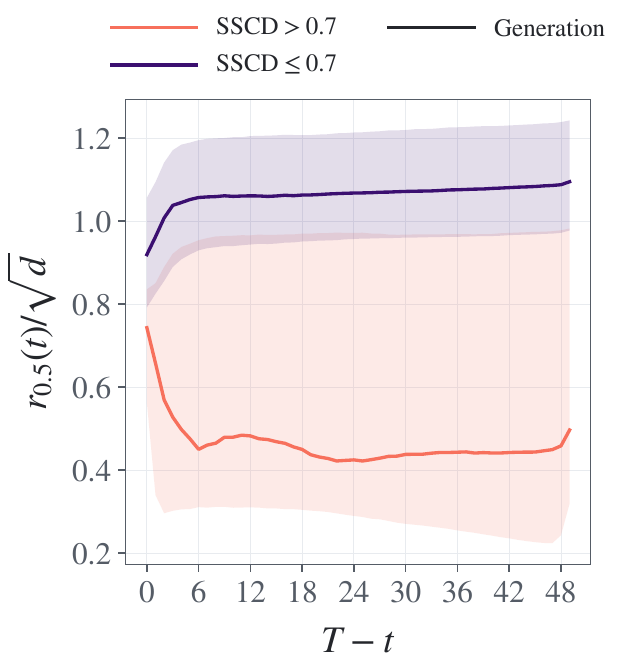}
        \caption{\autoref{prop:subsequent_conditional_reproduction}}
        \label{fig:sdv1_ddpm_prop4}
    \end{subfigure}\hfill
    \begin{subfigure}[t]{0.32\linewidth}
        \centering
        \includegraphics[
            width=\linewidth,
            viewport=0 0 357.78 334.464,
            clip
        ]{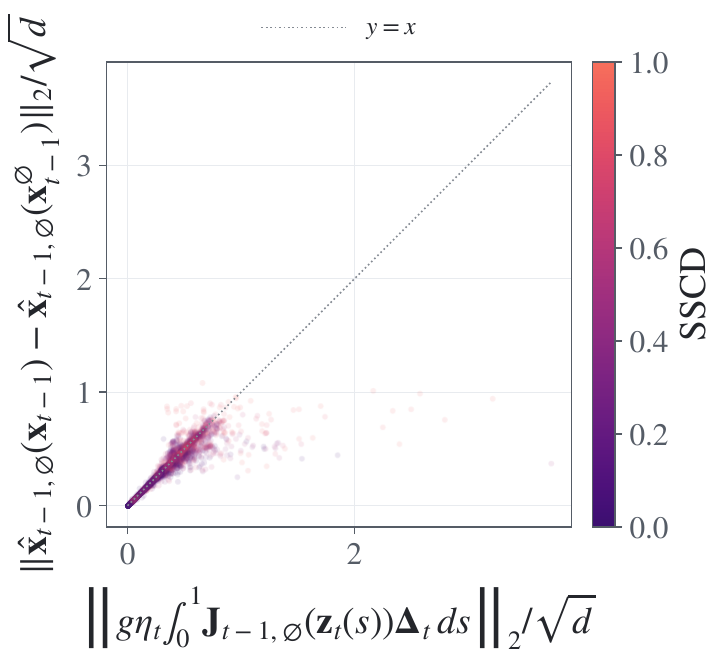}
        \caption{\autoref{thm:propagation_of_prompt_influence}}
        \label{fig:sdv1_ddpm_thm5}
    \end{subfigure}\hfill
    \begin{subfigure}[t]{0.32\linewidth}
        \centering
        \includegraphics[
            width=\linewidth,
            viewport=-2 -13 355.78 321.464,
            clip
        ]{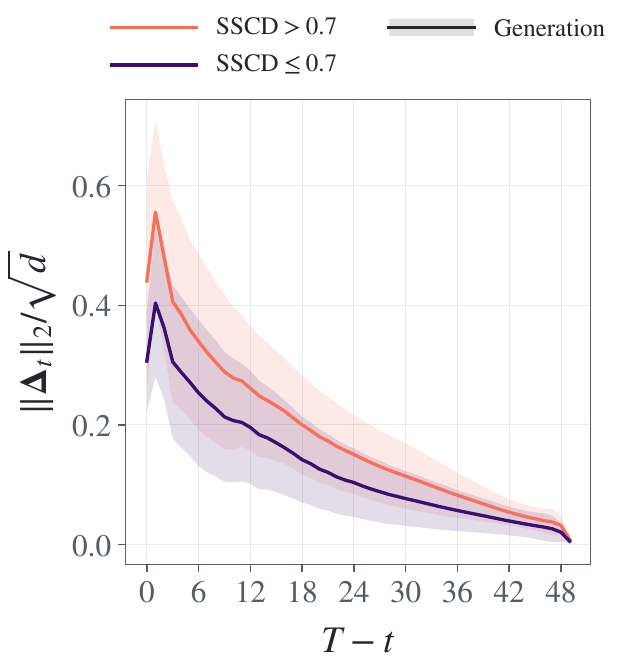}
        \caption{\autoref{thm:posterior_lag_dynamics}}
        \label{fig:sdv1_ddpm_thm8}
    \end{subfigure}

    \caption{
        \textbf{Denoising measurements on SDv1 (DDPM).}
        Coral and purple in (a,c,d,f) denote memorized and normal prompt groups, respectively (\autoref{sec:settings}).
        In (a), dashed and solid curves correspond to forward latents and Gaussian initializations, respectively.
        In (a--c,f), curves show means across prompts, and shaded bands indicate $\pm1$ standard deviation.
        Seeds are first averaged within each prompt in (b,f).
        In (d), curves and shaded bands show the median and interquartile range, respectively, with each prompt receiving equal total weight.
        In (e), point colors indicate each generated image's final SSCD, and the dotted diagonal denotes equality.
    }
    \label{fig:sdv1_ddpm}
\end{figure}

\begin{figure}[t]
    \centering
    \captionsetup[subfigure]{skip=2pt,hypcap=true}
    \setlength{\parskip}{0pt}

    \begin{subfigure}[t]{0.32\linewidth}
        \centering
        \includegraphics[
            width=\linewidth,
            viewport=-6 -7 351.78 327.464,
            clip
        ]{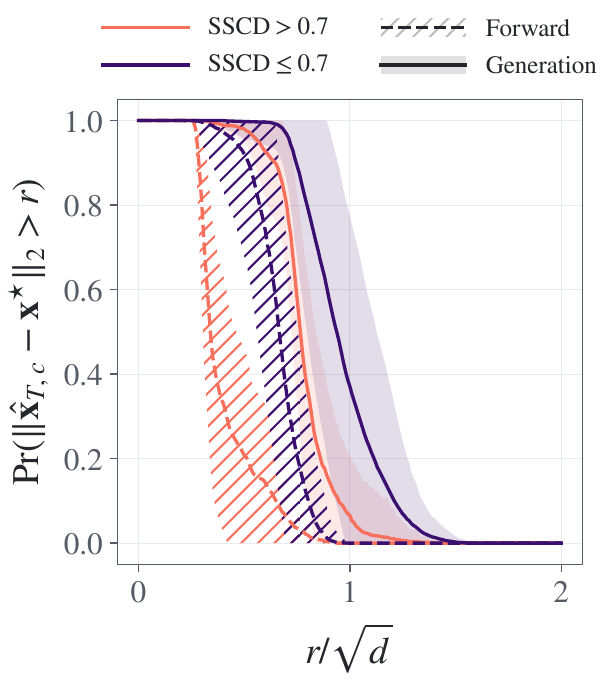}
        \caption{\autoref{lem:initial_conditional_reproduction}}
        \label{fig:sdv2_ddim_lem1}
    \end{subfigure}\hfill
    \begin{subfigure}[t]{0.32\linewidth}
        \centering
        \includegraphics[
            width=\linewidth,
            viewport=-2 -10.25 355.78 324.214,
            clip
        ]{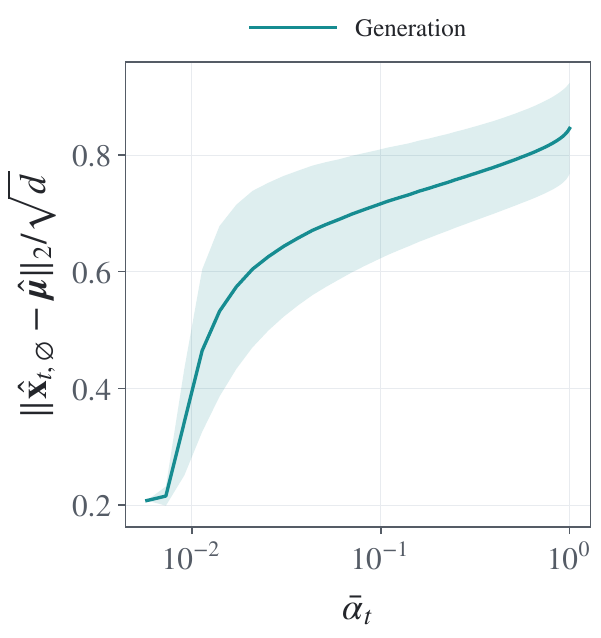}
        \caption{\autoref{lem:initial_unconditional_prediction}}
        \label{fig:sdv2_ddim_lem2}
    \end{subfigure}\hfill
    \begin{subfigure}[t]{0.32\linewidth}
        \centering
        \includegraphics[
            width=\linewidth,
            viewport=-5 -7 352.78 327.464,
            clip
        ]{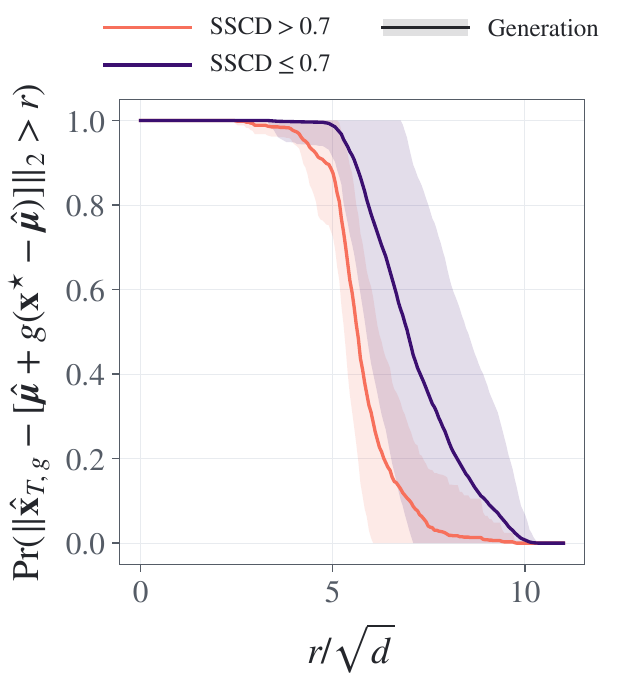}
        \caption{\autoref{thm:initial_cfg_amplification}}
        \label{fig:sdv2_ddim_thm3}
    \end{subfigure}

    \par\vspace{6pt}

    \begin{subfigure}[t]{0.32\linewidth}
        \centering
        \includegraphics[
            width=\linewidth,
            viewport=-2 -13 355.78 321.464,
            clip
        ]{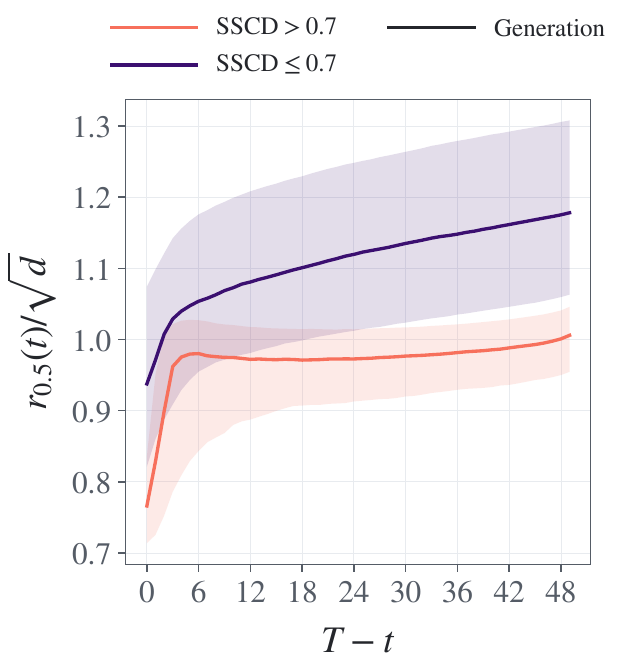}
        \caption{\autoref{prop:subsequent_conditional_reproduction}}
        \label{fig:sdv2_ddim_prop4}
    \end{subfigure}\hfill
    \begin{subfigure}[t]{0.32\linewidth}
        \centering
        \includegraphics[
            width=\linewidth,
            viewport=0 0 357.78 334.464,
            clip
        ]{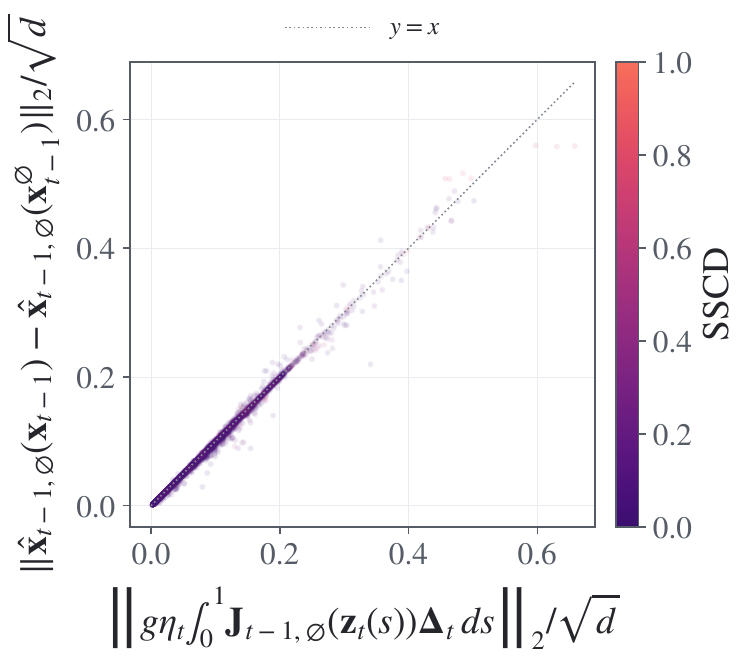}
        \caption{\autoref{thm:propagation_of_prompt_influence}}
        \label{fig:sdv2_ddim_thm5}
    \end{subfigure}\hfill
    \begin{subfigure}[t]{0.32\linewidth}
        \centering
        \includegraphics[
            width=\linewidth,
            viewport=-2 -13 355.78 321.464,
            clip
        ]{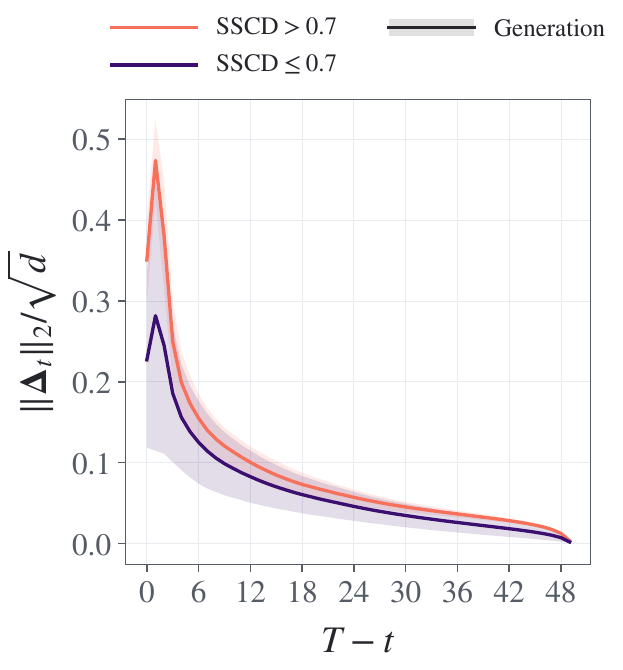}
        \caption{\autoref{thm:posterior_lag_dynamics}}
        \label{fig:sdv2_ddim_thm8}
    \end{subfigure}

    \caption{
        \textbf{Denoising measurements on SDv2 (DDIM).}
        Coral and purple in (a,c,d,f) denote memorized and normal prompt groups, respectively (\autoref{sec:settings}).
        In (a), dashed and solid curves correspond to forward latents and Gaussian initializations, respectively.
        In (a--c,f), curves show means across prompts, and shaded bands indicate $\pm1$ standard deviation.
        Seeds are first averaged within each prompt in (b,f).
        In (d), curves and shaded bands show the median and interquartile range, respectively, with each prompt receiving equal total weight.
        In (e), point colors indicate each generated image's final SSCD, and the dotted diagonal denotes equality.
    }
    \label{fig:sdv2_ddim}
\end{figure}

\begin{figure}[t]
    \centering
    \captionsetup[subfigure]{skip=2pt,hypcap=true}
    \setlength{\parskip}{0pt}

    \begin{subfigure}[t]{0.32\linewidth}
        \centering
        \includegraphics[
            width=\linewidth,
            viewport=-6 -7 351.78 327.464,
            clip
        ]{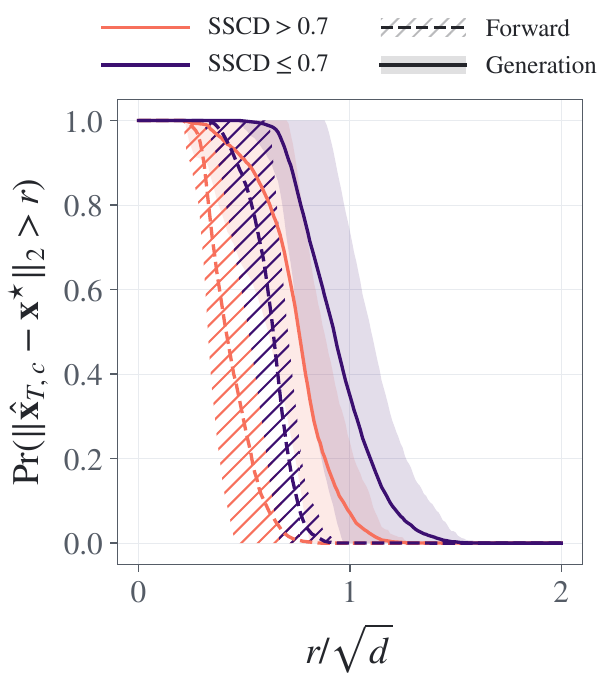}
        \caption{\autoref{lem:initial_conditional_reproduction}}
        \label{fig:realvis_ddim_lem1}
    \end{subfigure}\hfill
    \begin{subfigure}[t]{0.32\linewidth}
        \centering
        \includegraphics[
            width=\linewidth,
            viewport=-2 -10.25 355.78 324.214,
            clip
        ]{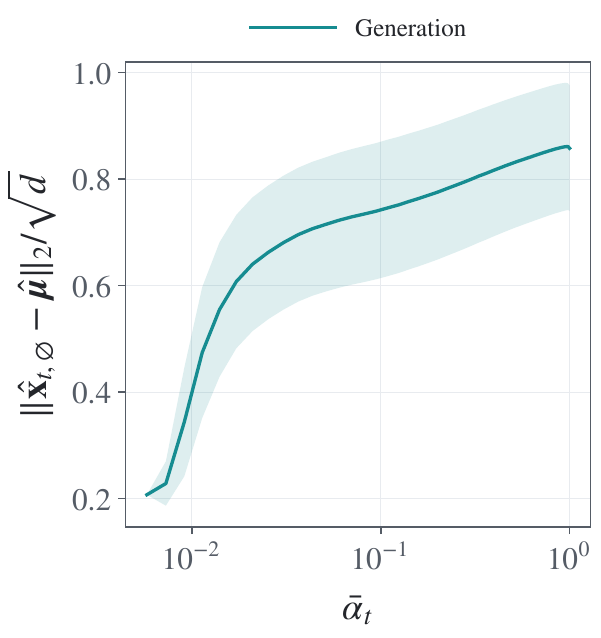}
        \caption{\autoref{lem:initial_unconditional_prediction}}
        \label{fig:realvis_ddim_lem2}
    \end{subfigure}\hfill
    \begin{subfigure}[t]{0.32\linewidth}
        \centering
        \includegraphics[
            width=\linewidth,
            viewport=-5 -7 352.78 327.464,
            clip
        ]{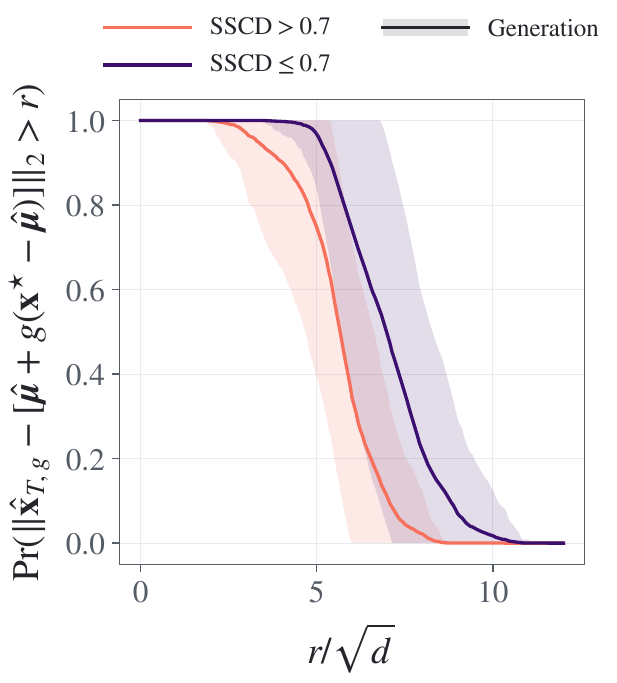}
        \caption{\autoref{thm:initial_cfg_amplification}}
        \label{fig:realvis_ddim_thm3}
    \end{subfigure}

    \par\vspace{6pt}

    \begin{subfigure}[t]{0.32\linewidth}
        \centering
        \includegraphics[
            width=\linewidth,
            viewport=-2 -13 355.78 321.464,
            clip
        ]{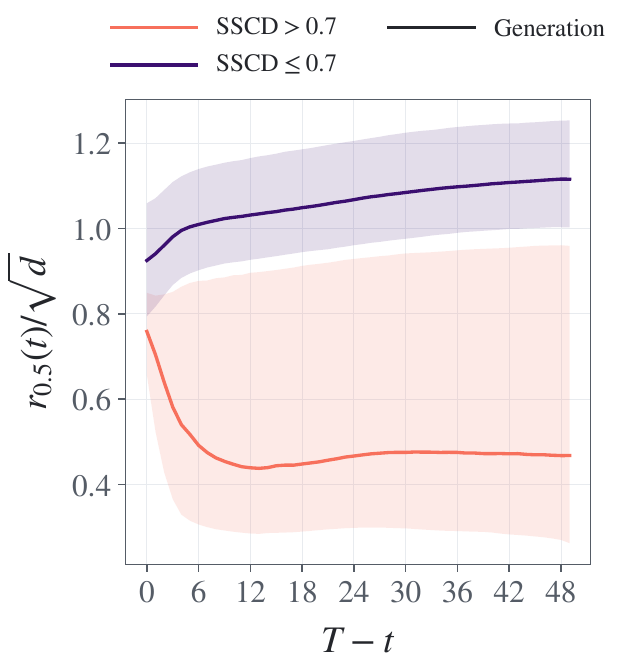}
        \caption{\autoref{prop:subsequent_conditional_reproduction}}
        \label{fig:realvis_ddim_prop4}
    \end{subfigure}\hfill
    \begin{subfigure}[t]{0.32\linewidth}
        \centering
        \includegraphics[
            width=\linewidth,
            viewport=0 0 357.78 334.464,
            clip
        ]{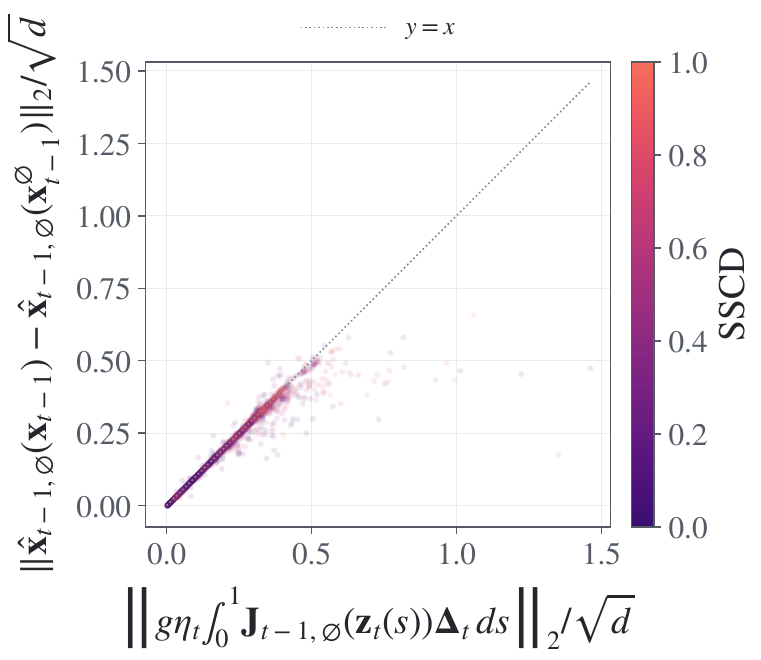}
        \caption{\autoref{thm:propagation_of_prompt_influence}}
        \label{fig:realvis_ddim_thm5}
    \end{subfigure}\hfill
    \begin{subfigure}[t]{0.32\linewidth}
        \centering
        \includegraphics[
            width=\linewidth,
            viewport=-2 -13 355.78 321.464,
            clip
        ]{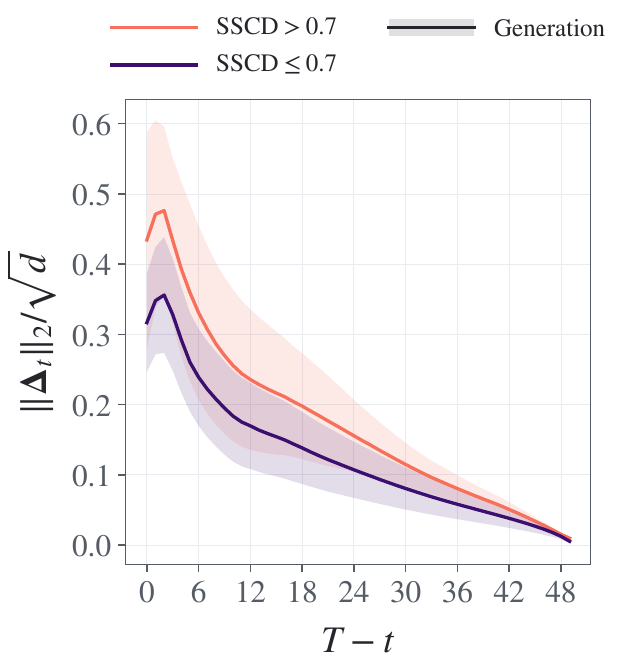}
        \caption{\autoref{thm:posterior_lag_dynamics}}
        \label{fig:realvis_ddim_thm8}
    \end{subfigure}

    \caption{
        \textbf{Denoising measurements on RealVis (DDIM).}
        Coral and purple in (a,c,d,f) denote memorized and normal prompt groups, respectively (\autoref{sec:settings}).
        In (a), dashed and solid curves correspond to forward latents and Gaussian initializations, respectively.
        In (a--c,f), curves show means across prompts, and shaded bands indicate $\pm1$ standard deviation.
        Seeds are first averaged within each prompt in (b,f).
        In (d), curves and shaded bands show the median and interquartile range, respectively, with each prompt receiving equal total weight.
        In (e), point colors indicate each generated image's final SSCD, and the dotted diagonal denotes equality.
    }
    \label{fig:realvis_ddim}
\end{figure}

\autoref{fig:sdv1_ddpm,fig:sdv2_ddim,fig:realvis_ddim} show results for SDv1 (DDPM), SDv2 (DDIM), and RealVis (DDIM), respectively.
The outcomes closely mirror those reported for SDv1 (DDIM) in the main text (\autoref{fig:sdv1_ddim}), supporting our findings across different diffusion models and samplers.

\clearpage
\section{Related work}
\label{app:related_work}

\textbf{Memorization and generalization.}
Prior work has studied the relationship between memorization and generalization in diffusion models~\citep{kadkhodaie2024generalization}.
\citet{vandenburg2021memorization} showed that removing a sample from training data can induce local density changes, which they use to quantify sample-level memorization.
Other studies found that duplicated images are more likely to be reproduced~\citep{somepalli2023forgery,somepalli2023copying,carlini2023extraction}.
Theoretical analyses also examine how empirical score optima, training dynamics, and regularization affect memorization~\citep{gu2025memorization,baptista2025regularization,bonnaire2025dynamical}.
Recent works propose geometric views: memorization is related to the local intrinsic dimensions of the data and learned manifolds~\citep{ross2025geometric}.
Similarly, \citet{jeon2025sharpness} link memorization to sharp regions of the log density.

\textbf{Text guidance and memorization.}
Prior work has established that text guidance is closely related to memorization.
For example, \citet{somepalli2023copying} highlight the role of text conditioning, and \citet{chen2024memorizationfree} propose Anti-Memorization Guidance (AMG) to mitigate memorization associated with duplicated images, duplicated captions, and highly specific prompts.
\citet{wen2024detecting} use the magnitude of the difference between conditional and unconditional noise predictions for detection and mitigation.
Other studies identify concentrated cross-attention on trigger tokens~\citep{ren2024crossattention} and cross-attention neurons that trigger replication~\citep{hintersdorf2024nemo}.
\citet{jain2025basin} provide an attraction-basin account and show how the timing of CFG affects memorization.

Our analysis complements these perspectives by explicitly deriving how the prompt's contribution to a CFG update changes the next unconditional clean prediction.
We connect this mechanism to conditional target reproduction and CFG amplification at the first denoising step.
We also bound the final reproduction error using both branches' target posterior probabilities at the final denoising step and show that closer agreement alone does not establish target reproduction.
\section{Limitations and future work}
\label{app:limitations_and_future_work}

Our analysis is currently limited to diffusion models and has not yet been extended to recent flow-based generative models.
Since flow models also generate samples through continuous trajectories, a similar analysis may reveal whether guidance amplifies and propagates information about the target through the same underlying mechanism.
Future work includes this extension and the development and comparison of mitigation methods that weaken conditional target reproduction, reduce early guidance-induced latent changes, or prevent subsequent unconditional target reproduction.



\end{document}